\documentclass[review,dvipsnames]{elsarticle}

\usepackage{lineno,hyperref,booktabs,longtable}
\usepackage[justification=raggedright]{caption}
\modulolinenumbers[5]
\usepackage[showframe=false]{geometry}
\usepackage{changepage}
\usepackage{fourier} 
\usepackage{array}
\usepackage{makecell}
\usepackage{listings}
\usepackage{xcolor}
\usepackage{todonotes}
\usepackage{multirow}
\usepackage{tikz}
\usepackage{natbib}

\usepackage{etoolbox}
\patchcmd{\pprintMaketitle}
 {\ifvoid\absbox\else\unvbox\absbox\par\vskip10pt\fi}
 {\ifvoid\absbox\else\clearpage\unvbox\absbox\par\vskip30pt\fi}
 {}{}
\patchcmd{\pprintMaketitle}
 {\hrule\vskip12pt}
 {}
 {}{}
\patchcmd{\pprintMaketitle}
 {\hrule\vskip12pt}
 {}
 {}{}
\appto{\pprintMaketitle}{\clearpage}

\biboptions{authoryear}

\def\checkmark{\tikz\fill[scale=0.4](0,.35) -- (.25,0) -- (1,.7) -- (.25,.15) -- cycle;} 

\begin{document}

\begin{frontmatter}

\title{Volume-Centred Range Bars: Novel Interpretable Representation of Financial Markets Designed for Machine Learning Applications}

\author[1]{Artur Sokolovsky\corref{cor1}%
}
\author[2]{Luca Arnaboldi}
\author[1]{Jaume Bacardit}
\author[1]{Thomas Gro\ss}

\address[1]{Newcastle University, School of Computing, 1 Science Square, Newcastle upon Tyne NE4 5TG, UK}

\address[2]{University of Edinburgh, School of Informatics, 10 Crichton St, Newington, Edinburgh EH8 9AB, UK}


\begin{abstract}
Financial markets are a source of non-stationary multidimensional time series which has been drawing attention for decades. Each financial instrument has its specific changing-over-time properties, making its analysis a complex task. Hence, improvement of understanding and development of more informative, generalisable market representations are essential for the successful operation in financial markets, including risk assessment, diversification, trading, and order execution.

In this study, we propose a volume-price-based market representation for making financial time series more suitable for machine learning pipelines. 
We use a statistical approach for evaluating the representation. Through the research questions, we investigate, i) whether the proposed representation allows the more efficient design of machine learning models; ii) whether the proposed representation leads to increased performance over the price levels market pattern; iii) whether the proposed representation performs better on the liquid markets, and iv) whether SHAP feature interactions are reliable to be used in the considered setting.

Our analysis shows that the proposed volume-based method allows successful classification of the financial time series patterns, and also leads to better classification performance than the price levels-based method, excelling specifically on more liquid financial instruments. 
Finally, we propose an approach for obtaining feature interactions directly from tree-based models and compare the outcomes to those of the SHAP method. This results in the significant similarity between the two methods, hence we claim that SHAP feature interactions are reliable to be used in the setting of financial markets.
\end{abstract}

\begin{keyword}
Applied ML \sep Volume Profiles \sep Boosting Trees \sep Explainable ML \sep Computational Finance
\end{keyword}

\end{frontmatter}

%

\section{Introduction}



Time series analysis rules for trading have historically been used with mixed successes~\cite{park2007we}.
This is in part due to the simplicity of the chosen rules~\cite{cervello2015stock}.
These rules are very often used in high frequency trading, where simple patterns are used to make thousands of trades with small marginal profits.
However, even studies based on more complex flag pattern~\cite{leigh2002analysis} rules, do so in a black box manner.
What we mean by this is that the authors use their own knowledge of known trading patterns of interest, and use machine learning to automatically discover these scenarios.
The performance of these has been shown to be particularly effective~\cite{cervello2015stock,leigh2002analysis}, disclaiming previous work discussing the inability for these approaches to work~\cite{ratner1999tests}.
We follow on with this work adapting a different type of pattern based analysis, however follow on with the usage of more interpretive machine learning methods further increasing potency of this approach for both a profitability and usability standpoint.

As automated machine learning (ML) algorithms take over several aspects of trading, accountability becomes an important factor at play~\citep{raji2020closing,ahmad2020fairness,martin2019ethical,liu2019beyond,wieringa2020account}. 
Due to the complexity, and unintuitive nature of most machine learning models, it is unreasonable to expect that users, who may or may not be an expert in the field, to understand how a model works and comes to decisions.
Furthermore, even if a user may understand exactly how the model works, following a decisions through a very complex and large set of mathematical steps, is unreasonable and often unfeasible for any real-world application~\citep{biran2017human}.
ML has historically been treated as a black box, data is input in the model, and a prediction is made.
However, whilst this kind of approach may very well trade with high levels of profitability, what happens when it does not?
The downside of the black box approach is that we are unable to understand why certain decisions are made - consequently when things go wrong, it becomes difficult to ascertain why~\citep{wieringa2020account,adadi2018peeking}.
What is instead ideal is to have explainable machine learning model. An explainable model, is the one which is non-inherently understandable by design, but whose results can be explained by means of post-hoc methods~\citep{adadi2018peeking}.
Through more explainable AI; with careful choice of algorithms and data curation we are able to ascertain why decisions are made, why things go wrong and consequently adjust our strategies to maintain profitability.

Explainable machine learning is an active research area across many different disciplines~\citep{wieringa2020account}; however we have yet to reach a consensus on how to achieve perfect understanding as several challenges arise~\citep{hoffman2018metrics,hagras2018toward,doran2017does}. 
It is generally understood, that focusing on more understandable machine learning algorithms, such as logistic regressions, and with careful feature selection we can greatly improve understanding.
Whilst this is a very active area in certain domains, such as medicine~\citep{holzinger2017we}, comparatively little research has been applied to finance and trading.
In this work we showcase a combination of novel state of the art machine learning techniques and statistics methods to create effective automated trading means that can both potentially garner profits whilst still being potentially understandable by a human trader to monitor and adjust as needed.

Generally speaking, one can achieve explainability in AI in three ways~\citep{adadi2018peeking}: 1) using more understandable algorithms, 2) reverse engineering the estimator to understand how it came to a decision, and/or 3) domain-specific adjustment of the input entries and feature design. 
Whilst the first approach is the more desirable of the two, as explanability comes inbuilt, there is often a trade-off between using simpler, more understandable models that may be less accurate, and more complex (less understandable) models that may well be highly accurate~\citep{breiman2001statistical}.
The second approach is gaining traction in recent years, focusing on using: i. visualisations, ii. natural language explanations and iii. explanations by example~\citep{adadi2018peeking}. 
The third approach leads to optimal input entries and feature space. Applied to financial markets, this means careful selection of the events (time series patterns in our case) as well as use of the most relevant features, leading to less convoluted model explanations.   
Which approach is most suitable depends on the prediction tasks. In this work we use domain knowledge to design the time series patterns of interest. 

\subsection{Research Gap}
The search of novel approaches to financial markets analysis is a common and a well-known problem of the domain, also referred as alpha mining~\citep{tulchinsky2019finding}. The term "alpha" is defined as a trading signal which potentially adds value to the financial portfolio. The trading signal may be represented as a rule of sampling points of interest from the time series. It can be based on domain knowledge or be completely abstract. The latter can be mined in an automatic way~\citep{zhang2020autoalpha}. One of the ways of mining alphas is through combining existing fundamental approaches like technical indicators or price action patterns~\citep{tulchinsky2019finding}. 
In the current work we propose a novel market pattern that is based on domain knowledge, execution volumes, and price action. Moreover, it is highly flexible and designed with machine learning applications in mind. Hence, the proposed pattern is not only useful on its own as an approach of market analysis, but also is potentially a valuable asset for alpha mining systems.

To our knowledge, there is no study investigating whether SHAP explanations are applicable to machine learning models aimed at pattern classification in financial markets. The main reason why it is essential to investigate this in the field of finance, is that the characteristics of these datasets are very unique. Namely, they have a high non-stationarity and a small probability of finding a persistent single feature with a strong effect size (caused by constantly evolving competition). One of the consequences is that the target model performance in the domain is often below the one expects in other fields~\citep{Dixon2017Classification-basedNetworks}. 
Since ML-supported trading systems rely on multiple features~\cite{de2018advances}, in the current study, we focus on feature interactions and aim to investigate the reliability of feature interactions extracted from machine learning models using SHAP. 

Considering the wide scope of the current study, for any terms related to finance or machine learning that may be unknown, we introduce a glossary in the supplementary material (Tab.~\ref{tab:glossary}) for the readers convenience.

\subsection{Contributions}
The contributions of the current study are three-fold. 
Firstly,we introduce a new type of a market pattern, called volume-centred range bars (VCRB), that is designed specifically for machine learning applications, and compare it to a commonly known price action pattern, price levels, which was extensively studied from perspectives of automatic detection and classification~\citep{sokolovsky2020machine}. 
Both patterns performance is analysed across two different financial instruments - CME Globex British Pound futures (B6 symbol) and S\&P E-mini Futures (ES symbol). In the current work, the two financial instruments serve as two datasets of different properties.

Secondly, we identify the optimal conditions for the proposed market pattern from the perspective of market liquidity.

Finally, we investigate whether commonly accepted explainability methods like SHAP can be applied in the financial markets setting. 
This is done by comparing SHAP feature interactions to the ones extracted explicitly from the model on example of CatBoost~\citep{Prokhorenkova2018Catboost:Features}. 
To perform the comparison, we propose a way of obtaining any order feature interactions directly from tree-boosting models (using CatBoost as an example), and then compare the results to the SHAP approach. 
CatBoost is a state-of-art boosting trees algorithm, it is explainable by post-hoc explanations i.e. SHAP~\citep{NIPS2017_8a20a862}, efficient and robust~\citep{Prokhorenkova2018Catboost:Features}.
To formally address the research gap, we make use of several statistical techniques, namely: effect sizes and hypothesis testing to measure the effects in a uniform and comparable way and assess significance of the results, backtesting~\footnote{Python Library for testing trading strategies available at: https://www.backtrader.com} to simulate trading performance, as well as SHAP~\citep{NIPS2017_8a20a862} and Monoforest-based approaches~\citep{NEURIPS2019_1b9a8060} to allow explanation of the decisions. 
All this allows us to assess the extent to which SHAP, as an indirect and commonly accepted method, can be used in the financial markets domain.



\subsection{Research Questions and Paper Outline}
In this study we are answering four research questions. The questions are aimed at addressing the stated research gap in detail. That includes comparison of the proposed method to a commonly known one, identification of the optimal operating conditions, as well as formal assessment of the interpretability.

The proposed method allows to extract samples of certain properties from the financial time series.
In order to make use of the extracted patterns, we classify them into scenarios as a way of predicting the future.
We consider two of the general price behaviours which can be traded - price crossing the target and price reversing from the target. In the current study, the target is either a price level (extremum) or PoC (Point of Control - the price with the largest volume in the volume profile).

Classification of the scenarios depends a lot on the feature space. 
By answering the first two research questions, we investigate whether the chosen market microstructure-based feature space and model are appropriate for the proposed method and further experiments. Below we list the research questions, while the hypotheses are formulated in the \nameref{methods} section. This way we gradually introduce the reader to the technical aspects of the study. 

\begin{center}
\textit{\textbf{RQ1:} Given our proposed volume-based pattern extraction method baseline performance (as always-positive), can we further increase it with a domain-led feature engineering and ML model?    }
\end{center}

\begin{center}
\textit{\textbf{RQ2:} Are the proposed volume-based patterns potentially better suitable for trading than price level-based? }
\end{center}
Aiming to answer this question as generally as feasible, we consciously avoid assessing the trading profitability for a specific trading strategy based no the two patterns.
Instead, we set up a classification task and compare the classification performance on them. 

For the third research question we hypothesise that volume-based patterns will perform statistically better on a more liquid market (S\&P E-mini in our case). 
The reasoning is that if there are more large players on the market, the price is less price-action driven or "noisy", hence the representation (pattern) involving volumes will be more complete. We use VCRB method for pattern extraction and expect it to be more suitable for liquid markets.

\begin{center}
\textit{\textbf{RQ3:} Does classification of volume-based patterns show better results on a liquid market? }
\end{center}
This final RQ arises from the field of explainable machine learning, seeking to assess to what extent we can rely on SHAP model explanations in the considered setting. 

\begin{center}
\textit{\textbf{RQ4:} Are feature interactions discovered with SHAP significantly associated with the ones obtained directly from the decision paths of the model? }
\end{center}
SHAP is a game theoretical model-agnostic framework for interpretations of the model outputs~\citep{NIPS2017_8a20a862}. 
Among other information, it provides a ranking of features based on contribution to the prediction per instance, and feature interactions.
Since SHAP provides an intuitive understanding and is easy to interpret, it is an ideal candidate to evaluate diverse real-world models.
Nevertheless, since SHAP values are an approximation of the original model, it would be useful to compare SHAP to an explicit model interpretation method.

The remainder of the paper is structured as follows: Section~\ref{methods} communicates the research protocol and evaluation criteria used for this paper; Section~\ref{sec:result} presents the results of our evaluation; Section~\ref{discussion} discusses the results and the extent to which our analysis is successful as well as future work; finally, Section~\ref{sec:conclusion} concludes the paper. Further analysis and extra details are provided in the supplementary materials after the bibliography.
We note the paper is structured so that each Research Question and Hypothesis is addressed and validated sequentially, with specific analysis addressing the question and providing evidence for or against the hypothesis.
\section{Material and Methods}
\label{methods}

In this section, we provide details of all the experiments performed, datasets used, model training and evaluation methodology.
Most of the discussion around the market structures will focus on the currently proposed volume bars, details of the price levels approach are discussed in the previous work~\citep{sokolovsky2020machine}.

\subsection{Datasets}
In the current study, we use a convenience sample of S\&P E-mini (ES) and British Pound (B6) futures instruments data, traded on CME Globex. The unseen data of an instrument is considered as a population.
Namely, we consider a time range of 39 months - from March 2017 until June 2020. When discussing the results, the data outside of this time range is considered a statistical population.

Since Order Book data is less commonly available and requires extra assumptions for pre-processing, we use tick data with only Time\&Sales per-tick statistics. 
Concretely: we obtain numbers of trades and volumes performed by aggressive sellers and buyers, at the bid and ask, respectively. Additionally, we collect millisecond-resolution time stamps on tick starts and ends as well as the prices of the ticks. 
In order to make the instruments comparable, we use price-adjusted volume-based rollover contracts data. Detailed data pre-processing is explained later on.

\subsection{Volume-centred range bars (VCRB)}
A volume profile is a representation of trading activity over a specified time and price range, its variations are also called market profiles.
They are commonly used for market characterisation (when considering daily volume profiles) and trading. 
Volume profiles are usually built from the temporal or price range bars, once every $n$ bars. 
These settings make the obtained profiles highly non-stationary and applying ML to them has the same drawbacks as feeding raw financial time series into a model. Namely, in such a setting models are hard to fit and require large datasets due to the constantly changing properties of the data.

\begin{figure}[!htb]
    \centering
    \includegraphics[width=1.0\textwidth]{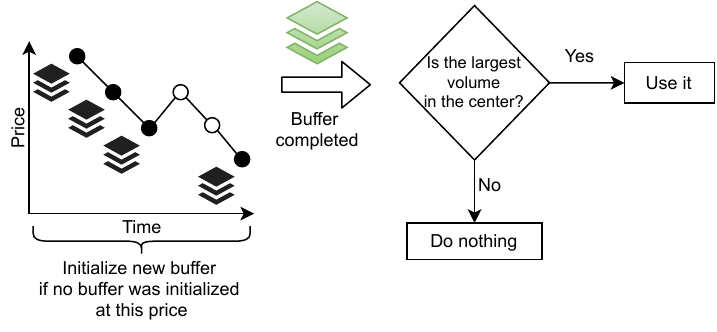}
    \caption{Visualisation of the volume-based pattern extraction approach. Entries filled with black indicate initializations of new buffers.}
    \label{fig:VCRBflow}
\end{figure}

In the proposed method we aim to increase the stationarity of the volume profiles, as well as obtain as many entries per time interval as possible. 
We illustrate a high-level diagram of the VCRB extraction pipeline in Figure~\ref{fig:VCRBflow}.
The core of the proposed method is a set of tick buffers simultaneously filled on a per-tick basis. A set is formed by new buffers started at a price if there is currently no incomplete buffer that has been initialised for the price. Consequently, there is a buffer initialised at every possible historical price. This way we ensure capturing all the samples potentially having the desired volume-centred configuration which is described in the next paragraph.
When the desired price range is reached for a buffer, it is considered complete and is not filled with the streamed ticks anymore - the same way as it works for conventional range bars.  
The price range is measured as a minimum tick price subtracted from the maximum price in the buffer. 

After the buffer is complete, we build its volume profile and check if the largest volume is in its centre. 
If so, we proceed with computing its features and labelling it. Otherwise, we ignore it. 
As we show later in ~\nameref{sec:result}, the tick buffer fulfilling the condition can be visualised as a volume-centred range bar (VCRB).
The largest volume in the volume profile of the buffer is called the Point of Control (PoC).
We use the PoC as a target for labelling, where price either reverses from or crosses it. 

From now on we refer to the price range is referred as \textit{range configuration}.

\subsection{Experiment Design}
In the current subsection, we describe all the components of the experiment design. We ensure that the design is as uniform as feasible across the experiments. When highlighting the differences, we refer to the experiments by the associated research questions - from RQ1 to RQ4.

\subsubsection{Label Design and Classification Setting}
For the sake of comparability, in the current study, we reuse the labelling procedure from the study by \citet{sokolovsky2020machine}. The labelling process starts when the VCRB is formed.
After the price reaches the target (POC or extremum of the pattern for VCRB and price level, respectively), there are two scenarios: it reverses or continues its movement. 
For the reversal, we require the price to move for at least 15 ticks from the target. 
For the crossings, we take the 3 ticks beyond the target. 
If the price crosses the target for less than 3 ticks and then reverses, we exclude this entry from training and label it as negative for testing. Based on the domain knowledge, we consider these numbers suitable for intraday trading scenarios - this is also supported by intraday standard deviations of the price. Optimisation of these numbers is out of the scope of the current study. 

Since price reversals require 15 ticks of price movement, they are directly applicable for trading and are considered positives in our binary classification tasks. Target price crossings are labelled based on 3 ticks price movements, hence their direct use in trading is limited and we consider them negatives. 
Consequently, price reversals are labelled as positives and crossings - as negatives.

We perform binary classification of the extracted VCRB and price levels patterns. 
Namely: we classify whether the entries are followed by price rebounds (reversals) from the target or target crossings. 
In the study, we use one of the novel tree boosting algorithms - CatBoost, which is known to be stable in wide ranges of model parameters and deal well with large feature spaces~\citep{Prokhorenkova2018Catboost:Features}. 

\subsubsection{Feature Space and Model Parameter Tuning}
We conduct two types of experiments - with and without feature selection and model parameter tuning. We have done so to study different aspects of the matter:
    \begin{itemize}
        \item Experiments with feature selection and model tuning. Since the process allows models with an optimal feature subset and model hyperparameters, these experiments are not limited by a particular feature set, but rather by an overall feature space. We use this setting for RQ1-3.
        \item Experiments with a fixed feature space and model parameters. These are aimed at studying feature interactions across datasets. Fixed feature space and model parameters ensure that any differences in the feature interactions are caused by the data or the model analysis approach and not by varying model configuration or feature space. 
        These are used specifically in RQ4.
    \end{itemize}
    
When designing the feature space we follow the same approach as suggested in \citet{sokolovsky2020machine} - we extract a set of features from the volume-based pattern or a price level - called Pattern features, and the second one from the most recent ticks before the target is approached (being 2 ticks away from it) - called Market Shift (MS) features. While the approach is preserved, the set of features is different due to a simplified dataset, where the Order Book data is not available. The feature space was designed with stationarity, generalisability, as well as domain knowledge in mind~\cite{de2018advances}.

In order to decrease the number of uncontrolled factors and their impact on the experiment design, we limit the feature space to only market microstructure-based features. 
We list the features and the associated equations in Table \ref{tab:featSpace}.
The provided equations are valid for volume-based patterns. 
Since the data is available only below or above the extrema in the case of the price level patterns, the features are computed slightly differently. Negative \textit{t} values are considered as an odd number of ticks distances from extrema and positive - as even numbers. We believe that this alteration is the closest possible to the original while preserving the domain knowledge at the same time. Numbers 237 and 21 in Table~\ref{tab:featSpace} were taken arbitrarily with three requirements: not round, within the trading time interval range, substantially different. We performed no optimisation of these parameters.

\begin{table}[!htbp]
\small
\caption{Features (referred to by code '[code]') used in the study in two stages, Stage 1 - Pattern and Stage 2 - Market Shift (MS)}
\centering   
\begin{tabular}{l|l}

     \multicolumn{2}{l}{
    \begin{tabular}{|c|ll|}
    \hline
    \multirow{ 15}{*}{\centering\rotatebox{90}{\textbf{Pattern features}}}& \textbf{Equation} & \textbf{Description} \\
     \hline
       & $\frac{\sum_{t \in [1;5]}^{p=X+t}(V_b)}{\sum_{t \in [-5;-1]}^{p=X+t}(V_b)}$ & Sum of upper (above PoC) bid volumes divided by lower ones [P0]\\[9pt]
        & $\frac{\sum_{t \in [1;5]}^{p=X+t}(V_a)}{\sum_{t \in [-5;-1]}^{p=X+t}(V_a)}$ & Sum of upper ask volumes divided by lower ones [P1] \\[9pt]
        & $\frac{\sum_{t \in [1;5]}^{p=X+t}(T_a)}{\sum_{t \in [-5;-1]}^{p=X+t}(T_a)}$ & Number of upper bid trades divided by lower ones [P2] \\[9pt]
        & $\frac{\sum_{t \in [1;5]}^{p=X+t}(T_b)}{\sum_{t \in [-5;-1]}^{p=X+t}(T_b)}$ & Number of upper ask trades divided by lower ones [P3] \\[9pt]
        & $\frac{\sum_{t \in [1;5]}^{p=X+t}(V_b)}{\sum_{t \in [1;5]}^{p=X+t}1}$ & Average upper bid trade size [P4] \\[9pt]
        & $\frac{\sum_{t \in [1;5]}^{p=X+t}(V_a)}{\sum_{t \in [1;5]}^{p=X+t}1}$ & Average upper ask trade size [P5] \\[9pt]
        & $\frac{\sum_{t \in [-5;-1]}^{p=X+t}(V_b)}{\sum_{t \in [-5;-1]}^{p=X+t}1}$ & Average lower bid trade size [P6] \\[9pt]
        & $\frac{\sum_{t \in [-5;-1]}^{p=X+t}(V_a)}{\sum_{t \in [-5;-1]}^{p=X+t}1}$, & Average lower ask trade size [P7] \\[9pt]
        & $\frac{\sum_{}^{p=X}V_b}{\sum_{t \in [1;5]}^{p=X+t}V_b}$ & Sum of PoC bid volumes divided by sum of upper bid volumes [P8] \\[9pt]
        & $\frac{\sum_{}^{p=X}V_a}{\sum_{t \in [1;5]}^{p=X+t}V_a}$ & Sum of PoC ask volumes divided by sum of upper ask volumes [P9] \\[9pt]
        & $\frac{\sum_{}^{p=X}V_b}{\sum_{t \in [-5;-1]}^{p=X+t}V_b}$ & Sum of PoC bid volumes divided by sum of lower bid volumes [P10] \\[9pt]
        & $\frac{\sum_{}^{p=X}V_a}{\sum_{t \in [-5;-1]}^{p=X+t}V_a}$ & Sum of PoC ask volumes divided by sum of lower ask volumes [P11] \\
        
       &  $\frac{\sum_{t}^{p=X+t}V_b}{\sum_{t}^{p=X+t}V_a}$ & \makecell[l]{Sum of bid volumes divided by sum of ask volumes as price X+t; \\ $t \in [-1;1]$ as different features [P12]} \\
       &  $\frac{\sum_{t}^{p=X+t}T_b}{\sum_{t}^{p=X+t}T_a}$ & \makecell[l]{Number of bid trades divided by number of ask trades as price X+t; \\ $t \in [-1;1]$ as different features [P13]} \\
        &  - & Side - below or above the price when the pattern is formed [P14] \\
    \hline
    \multirow{ 4}{*}{\centering\rotatebox{90}{\textbf{MS features}}}& $\frac{ \sum_{t,b}^{w=237}(V_b)}{\sum_{t,a}^{w=237}(V_a)}$ & Fraction of bid over ask volume for last 237 ticks [MS0] \\[6pt]
    & $\frac{ \sum_{t,b}^{w=237}(T_b)}{\sum_{t,a}^{w=237}(T_a)}$ &  Fraction of bid over ask trades for last 237 ticks [MS1] \\[6pt]
    & $\frac{ \sum_{t}^{w=237}V_b}{\sum_{t}^{w=237}V_a} - \frac{\sum_{t}^{w=21}V_b}{\sum_{t}^{w=21}V_a}$ & Fraction of bid/ask volumes for long  minus short periods [MS2] \\[6pt]
   & $\frac{ \sum_{t}^{w=237}T_b}{\sum_{t}^{w=237}T_a} - \frac{\sum_{t}^{w=21}T_b}{\sum_{t}^{w=21}T_a}$ & Fraction of bid/ask trades for long minus short periods [MS3] \\
   \hline
 \multirow{ 1}{*}{\centering\rotatebox{90}{\textbf{Key}}}
 & \multicolumn{2}{l|}{
        \begin{tabular}{lllll}
           t - number of ticks & a - ask & p - price & w - tick window & $P_{N}$ - neighbours until distance N\\
           V - volume & b - bid & T - trades & X - PoC or extremum &
        \end{tabular}
        }\\
    \hline
    \end{tabular}
}\\

\end{tabular}

    \label{tab:featSpace}
\end{table}

To perform the feature selection, we run a Recursive Feature Elimination with Cross-Validation (RFECV) process with a step of 1 and use the internal CatBoost feature importance measure for the feature ranking.
Model parameter tuning is done for the following model parameters, which are recommended for optimisation: i) the number of iterations; ii) maximum tree depth; iii) whether the temporal component of the data is considered; iv) L2-regularisation. We considered it infeasible to optimise over a larger number of parameters, as with the current setting the experiments take in total $>$7k (single-core) CPU hours.
When optimising the feature space and tuning the model parameters, the precision metric is used - the one directly related to the success of the model in a trading setting, as we explain later. 
\subsubsection{Performance Metrics}
For the RQ1 experiments, the performance is evaluated using the precision metric as it directly defines the potential profitability of the model. Namely, as the fraction of true positives over a sum of true positives and false positives, which defines the fraction of times the participant enters the market with a positive outcome. 
We do not account for negatives as they have no impact on the outcome, due to our approach of not trading these. 
We elaborate more on this in the~\nameref{discussion} section. 

We use binary Precision-Recall Area Under Curve  (PR-AUC) score as the metric for RQ2 and RQ3 as there we are comparing classification performance of differently imbalanced datasets and this measure is advised for use in this setting \citep{Saito2015TheDatasets}. 
For the sake of completeness and easier interpretation of the results, we provide the mentioned metrics together with ROC-AUC and f1-score for all the experiments. 
As there is no notion of performance in the experiments associated with RQ4, we describe its metric in Section: \nameref{sec:modelInterpret}.

\subsubsection{Model Evaluation}
Price-adjusted volume-based rollover pre-processing of the data allows us to split the data into 3-month chunks without any extra care for contract rollover. We do so since the length of contracts is different for ES and B6 and contracts cannot be compared directly. When training the models, we apply a temporal sliding window approach, using batch $N$ for training, $N+1$ for testing, for $N\in[1, B-1]$, where $B$ is the total number of 3-month batches available.
For the feature selection and model parameter tuning, we use 3-fold time-series cross-validation within the training batch with the final re-training on the whole batch. To support the readers, we summarise the experiment design for all the experiments in Table~\ref{tab:expDesign}. In the following subsections, we address the statistical component of the methodology.

\begin{table}[!htb]
    \caption{Experiment design across the experiments. F$_{sel}$ \& M$_{tune}$ column accounts for feature selection and model tuning. ``Comparison H$_{0}$/H$_{1}$'' column indicates the settings used to obtain null and alternative hypotheses data. ``Other'' indicates whether the test is conducted on both instruments (ES, B6) and whether only VCRB pattern extraction method was used. ``Metric'' corresponds to the measurement variable. All hypotheses are evaluated using Wilcoxon test.}
    \centering
    \begin{tabular}{c|c|c|c|c}
        Experiment & F$_{sel}$ \& M$_{tune}$ & Metric & Comparison H$_{0}$/H$_{1}$ & Other \\
         \hline
         RQ1 & \checkmark & Precision &No-info. / CatBoost & VCRB; ES, B6 \\
         RQ2 & \checkmark & PR-AUC & Price levels / VCRB & ES, B6 \\
         RQ3 & \checkmark & PR-AUC & B6 / ES  & VCRB \\

         RQ4 & - & Footrule distance & \makecell[c]{Bootstrap / SHAP, \\Monoforest-based}  & ES, B6 \\
    \end{tabular}

    \label{tab:expDesign}
\end{table}

\subsubsection{Statistical Evaluation}
In the current study, we use effect sizes following the formalism of comparison of treatment and control groups. For the two groups, we perform a pair-matched comparison choosing suitable measurement variables for each case, as shown in Table~\ref{tab:expDesign}. In our paper, the treatment and control groups are represented by the methods applied to the time series batches. 

Before evaluating the hypotheses, we compute Hedge's \textit{g$_{av}$} effect sizes for paired data together with the .95 confidence intervals (CIs). 
Communicating effect sizes makes it easier to interpret and compare results across studies, CIs allow generalising the results to the population (unseen data)~\citep{Lakens2013CalculatingANOVAs}. 
The effect sizes are significant if the CI does not overlap with the 0.0 effect size threshold. By definition, we expect the significant effect sizes to be present in the unseen data of the same financial instrument.

In order to formally reject the null hypotheses, we have to choose a suitable paired difference statistical test. Since the number of data batches is small, we take a conservative approach and require the selected test to be applicable to non-normally distributed data.
We choose Wilcoxon signed-rank test as the best suitable candidate~\citep{Wilcoxon1945IndividualMethods} and apply its single-sided version to validate the hypotheses of the study. 
We refer to null and alternative hypotheses as H$_{0X}$ and H$_{1X}$, respectively, with $X$ being the number of the research question.

Throughout the study, we set the significance level for the statistical tests of the study to $\alpha=.05$. 
Finally, we apply Bonferroni corrections to all the statistical tests within each experiment family ~\citep{ce1936teoria}. The experiment families are defined based on the data and objectives - each research question forms a separate experiment family.

\subsubsection{Model Interpretation}
The challenge of interpreting the modern ML models roots in their complexity. 
Even considering decision tree-based models, interpretation becomes problematic when the number of features grows. 
When it comes to interpretations, the number of decision paths of the CatBoost model is usually well beyond the limit of manual analysis. 
The recently proposed Monoforest~\citep{NEURIPS2019_1b9a8060} approach represents the tree ensemble as a set of polynomials and makes the decision paths uniform. Additionally, its implementation gives access to machine-readable decision paths. This allows us to retrieve the following information from each polynomial: the subset of the involved features, feature thresholds, as well as support ($w$) and contribution ($c$) to the output of the model. 

We assume that interactions between features take place if they are found in the same decision path. Please note that the way SHAP defines feature interactions is principally different, as we elaborate in \nameref{discussion} section.
In order to get interactions for the whole model, we average across the decision paths in the following way:
\begin{equation}
    I_{(F1,F2)} = \frac{ \sum_{i=0}^N c \times w}{N},
\end{equation}
where $N$ is the total number of decision paths containing features 1 and 2 ($F1$, $F2$), $c$ is the contribution of the decision path to the model output and $w$ is the support, computed as a number of times the path was activated in the training set divided by the total number of training entries.

If there is a single feature in the decision path, we consider it as a main effect. 
We include main effects into the interaction matrix in the same fashion as it is done for SHAP~\citep{NIPS2017_8a20a862} - as diagonal elements of the matrix. 
When there are more than two features in the decision path, we treat them as multiple pair-wise interactions to be able to represent them in a single 2d matrix and compare them to SHAP interactions. 
By considering decision paths with a fixed number of features, one can get interactions of a particular order. 
Moreover, these interactions are directly comparable across orders, models and datasets. 

\subsubsection{Backtesting}
\begin{figure}[!htb]
    \centering
    \includegraphics{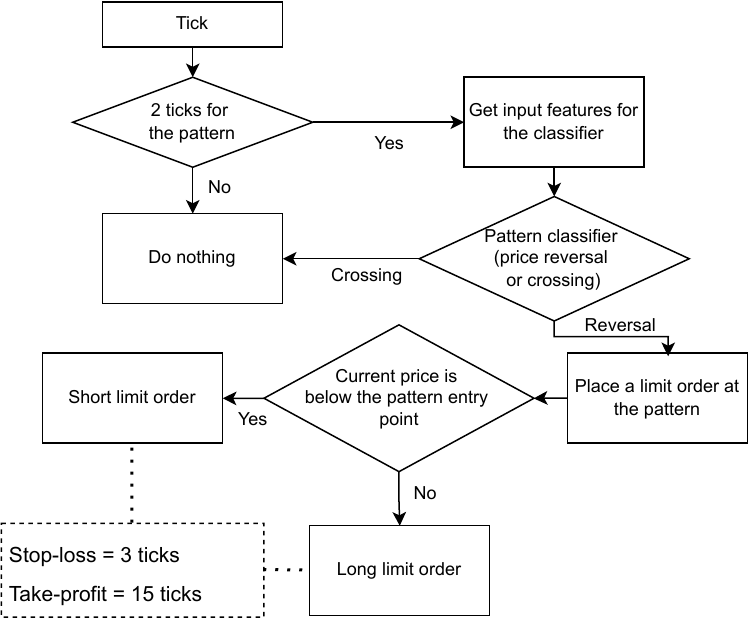}
    \caption{A diagram of the trading strategy used throughout the study.}
    \label{fig:strata}
\end{figure}

We perform backtesting of the proposed method in the Python Backtrader platform. 
We use the same strategy, assumptions and trading fees as used in the price levels study~\citep{sokolovsky2020machine}. The strategy is illustrated in Figure~\ref{fig:strata}. It takes the last tick as an input, then it checks whether the current price is 2 ticks away from the pattern entry point. The 2 ticks condition ensures that we have enough time to place the order, hence mitigating the risks of obtaining over-optimistic performance by modelling orders that cannot be executed in reality. 
When we are 2 ticks away from the entry point, we obtain the features and classify the entry as either price reversal or crossing. The strategy is aimed at trading only price reversals (flat market state), hence if the classification outcome is "crossing", we do nothing. Finally, if the price reversal is expected, depending on the current price with respect to the entry point, we place either a long or short limit order with the stated take-profit and stop-loss. 

This is a transparent and easy-to-grasp strategy. It is not only easy to implement, debug, and analyse, but also can be used in both semi-manual machine learning-aided and fully automated trading settings. 
Since the study is aimed at answering particular research questions and an absolute trading performance maximisation is not one of them, we make the best effort to keep the experiment design simple. The strategy is of similar nature and complexity as proposed in~\citep{cervello2015stock}. 

In the study we focus on performance comparison based on classification tasks over the backtesting simulations for two reasons: i) different trading intensities lead to varying impacts of the modelling assumptions; ii) limiting our comparison to a particular strategy significantly decreases the generality of the finding.
Elaborating on the first point: influences of order queues, slippages and bid-ask spreads are partially taken into account, however, it is obvious that these effects have increasing impacts with an increase in the trading frequency. 
Quantification of these is out of the scope of this study, however, might be very useful. We describe how the modelling assumptions might affect the backtesting results in the \nameref{discussion} section.
As a complementary analysis, in the current study, we report annual rolling Sharpe Ratios for all the range configurations, and for the price level method. 

Sharpe Ratio is a measure for assessing the performance of an investment or trading approach and can be computed as the following:
\begin{equation}
    S = \frac{R_p-R_f}{\sigma_p},
\end{equation}
where $R_p$ \& $R_f$ correspond to the portfolio and risk free returns, respectively, and $\sigma_p$ is a standard deviation of the portfolio return. While the equation gives a good intuition of the measure, in practice its annualised version is often computed. It assumes that daily returns follow the Wiener process distribution, hence to obtain annualised values, the daily Sharpe Ratio values are multiplied by $\sqrt{252}$ - the annual number of trading days. It should be noted that such an approach might overestimate the resulting Sharpe Ratios as returns auto-correlations might be present, violating the original assumption~\citep{Lo2002TheRatios}. In the study, we take a conservative approach and set the risk-free returns to 5\%. This is one of the measures aimed to address the issue of over-optimistic backtesting results known in finance~\citep{arnott2022s}.

\subsection{Addressing Research Questions}
In the current subsection, we list the conducted experiments associated with the research questions, as well as highlight any experiments-specific choices.

\subsubsection{VCRB method, CatBoost versus no-information estimator (RQ1)}
Firstly we assess the performance of the no-information and CatBoost classifiers and compare them. 
Prior to evaluating the hypotheses, we compute the effect sizes.
Then, we run the statistical test with the following hypotheses:

$H_{01}$: CatBoost estimator performs equally or worse than the no-information model. 

$H_{11}$: CatBoost estimator performs better than the no-information model. 

In the results section we report statistical test outcomes for the configuration with the largest effect sizes, the other configurations are reported in the supplementary materials. 

\subsubsection{VCRB vs price levels approach (RQ2)}
To answer the second research question we need to compare the performance of the two methods - price levels and VCRB. We obtain the PR-AUC classification performance for both methods and both instruments. Then, we compute the effect sizes, and, finally, we run the statistical test with the following hypotheses:

$H_{02}$: Price level patterns are classified with performance equally good or better than volume-based patterns. 

$H_{12}$: Volume-based patterns are classified with statistically better performance.

\subsubsection{VCRB method, ES versus B6 datasets (RQ3)}
Answering the third research question, we compare CatBoost classification performance on the VCRB-extracted data over the two datasets - B6 and ES, where the latter is far more liquid. After the PR-AUC classification performance is obtained, we compute the effect sizes and run the statistical tests with the following hypotheses: 

$H_{03}$: Both instruments perform comparably or performance on B6 is statistically better. 

$H_{13}$: Volume-based patterns are classified with statistically better performance for the instrument with higher liquidity (ES). 

\subsubsection{Relatedness of feature interactions from SHAP and decision paths (RQ4)}
\label{sec:modelInterpret}
To answer the last research question we need to get the data representing the null hypothesis - having no relation (potentially in contrast to SHAP and decision paths) and propose a method for assessing the relatedness. 
As the first step, we obtain feature interactions in a form of a square matrix using SHAP and the Monoforest-based method. To generate the null hypothesis data, we bootstrap (randomly sample with replacement) the interaction matrices' elements separately for both approaches. We choose to generate 500 bootstrapped entries for each method. As a result, we get two sets of matrices with the same value distributions as the original feature interaction matrices (SHAP and Monoforest-based). Since we performed bootstrapping, by definition there is no association between any two matrices from the two sets.

Since both methods output interactions scaled differently, we need to make the matrices comparable. To do so, we rank the interaction strengths within each matrix. 
After that, we compare the ranks across the two methods by computing their Footrule distances~\citep{Spearman1904TheThings}. This measure is designed specifically for ranks data and computed as the following:
\begin{equation}
    D = |R^A_{1}-R^B_{1}|+|R^A_{2}-R^B_{2}|+...+|R^A_{n}-R^B_{n}|,
\end{equation}
where $R$ is the position order of the element in the rankings $A$ \& $B$ of the length $n$, and the lower indices correspond to the compared elements.
 Later the distances are used as a proxy to assess the relatedness of the methods - the larger the distance, the weaker the relationship. 

In order to get a reliable null hypothesis distance, we compute the mean of the distances between the bootstrapped matrices. At this point, the relatedness of SHAP and decision paths methods can be compared against the bootstrapped data. 
To quantify the differences, we obtain the effect sizes on the Footrule distance (instead of the classification performance used in the other RQs). While it is advised to use Glass's $\Delta$ in case of the significantly different standard deviations~\citep{Lakens2013CalculatingANOVAs} between the groups, we do not follow this advice. Glass's $\Delta$ normalises the effect size by the standard deviation of the control group, which is the bootstrapped data in our case. Its standard deviations are expected to be much smaller than the actual data, leading to an abnormal increase in the effect sizes. Moreover, there is no Glass's $\Delta$ adapted for the paired data. Hence, the reader should keep in mind that the communicated effect sizes for the RQ4 might be overestimated. 
Finally, we run the statistical test with the following hypotheses:

$H_{04}$: There is no difference between Footrule distances on ranks of SHAP-decision paths and bootstrapped feature interactions matrices or SHAP-decision paths are larger. 

$H_{14}$: There is a difference between Footrule distances on ranks of SHAP-decision paths and bootstrapped feature interactions matrices with SHAP-decision paths being smaller than bootstrapped.

\section{Results}
\label{sec:result}

This section presents all the results from our methodology described in the previous section,
We reserve a detailed analysis of how the results match our hypothesis for Sec.~\ref{discussion}, and this section only contains the outcomes of the experiments.

\subsection{Pattern extraction from the datasets}
In this subsection, we report statistics on the original datasets as well as numbers of entries obtained from every dataset batch, for both extraction methods and financial instruments.

In Table~\ref{tab:data} we show the numbers of ticks and total volumes per batch for both instruments.
\begin{table}[!htb]
    \caption{Original datasets statistics. Volume columns correspond to the total volume traded per the stated time interval. The Ticks columns show the numbers of ticks per the time interval.}
    \centering
    \begin{tabular}{l|cc|cc}
        {Batch} & \multicolumn{2}{c|}{\textbf{ES}} & \multicolumn{2}{c}{\textbf{B6}} \\
        \hline
        {} &  Volume &  Ticks &  Volume &  Ticks \\
        \hline
        3/17 to 6/17  &   86123932 &    151965 &    6663094 &    118098 \\
        6/17 to 9/17  &   82384394 &    132964 &    6575559 &    115576 \\
        9/17 to 12/17 &   73925568 &     98600 &    7971963 &    142548 \\
        12/17 to 3/18 &   88050918 &    517451 &    7588515 &    159565 \\
        3/18 to 6/18  &   96653879 &    536280 &    7267680 &    131215 \\
        6/18 to 9/18  &   71968775 &    235841 &    6956995 &    116275 \\
        9/18 to 12/18 &  109410969 &    581345 &    7128825 &    155895 \\
        12/18 to 3/19 &   95948559 &    673838 &    6078533 &    136554 \\
        3/19 to 6/19  &   92201997 &    378788 &    6643282 &    132441 \\
        6/19 to 9/19  &   93229922 &    458141 &    5677468 &     94248 \\
        9/19 to 12/19 &   75613694 &    343666 &    7193235 &    153081 \\
        12/19 to 3/20 &  101547199 &    587658 &    6122643 &    100381 \\
        3/20 to 6/20  &  126756329 &   3482845 &    5593815 &    270096 \\
    \end{tabular}

    \label{tab:data}
\end{table}

We see in Table~\ref{tab:extrData} that in the year 2017 there are more entries for VCRB B6 than for ES, later the situation reverses. 
Overall, there are around 5-8 times fewer entries for the price level-based method in comparison to the volume-based. We address the potential consequences of these differences in the~\nameref{discussion} section.

\begin{table}[!htb]
    \caption{Numbers of extracted patterns for volume-based (VCRB) range 7, and price level-based (PL) methods, reported for the analysed data sets, both instruments.}
    \centering
        \begin{tabular}{l|cc|cc}
            {Batch} & \multicolumn{2}{c|}{\textbf{VCRB}} & \multicolumn{2}{c}{\textbf{PL}} \\
            \hline
            {} & ES & B6 & ES & B6 \\
            \hline
            3/17 to 6/17  &   2236 &  2785 &    458 &    278 \\
            6/17 to 9/17  &   1998 &  2789 &    360 &    267 \\
            9/17 to 12/17 &   1416 &  3440 &    268 &    378 \\
            12/17 to 3/18 &  10471 &  4018 &   1643 &    418 \\
            3/18 to 6/18  &  10231 &  3125 &   1695 &    335 \\
            6/18 to 9/18  &   4182 &  2643 &    736 &    287 \\
            9/18 to 12/18 &  11876 &  3789 &   1926 &    370 \\
            12/18 to 3/19 &  13937 &  3354 &   2234 &    348 \\
            3/19 to 6/19  &   6940 &  3480 &   1186 &    358 \\
            6/19 to 9/19  &   9111 &  2329 &   1413 &    235 \\
            9/19 to 12/19 &   6202 &  3878 &   1060 &    410 \\
            12/19 to 3/20 &  12886 &  2505 &   1856 &    227 \\
            3/20 to 6/20  &  88486 &  8373 &  12085 &    738 \\
        \end{tabular}

    \label{tab:extrData}
\end{table}

\subsection{Volume-centred range bars}
\begin{figure}
    \centering
    \includegraphics[width=\textwidth]{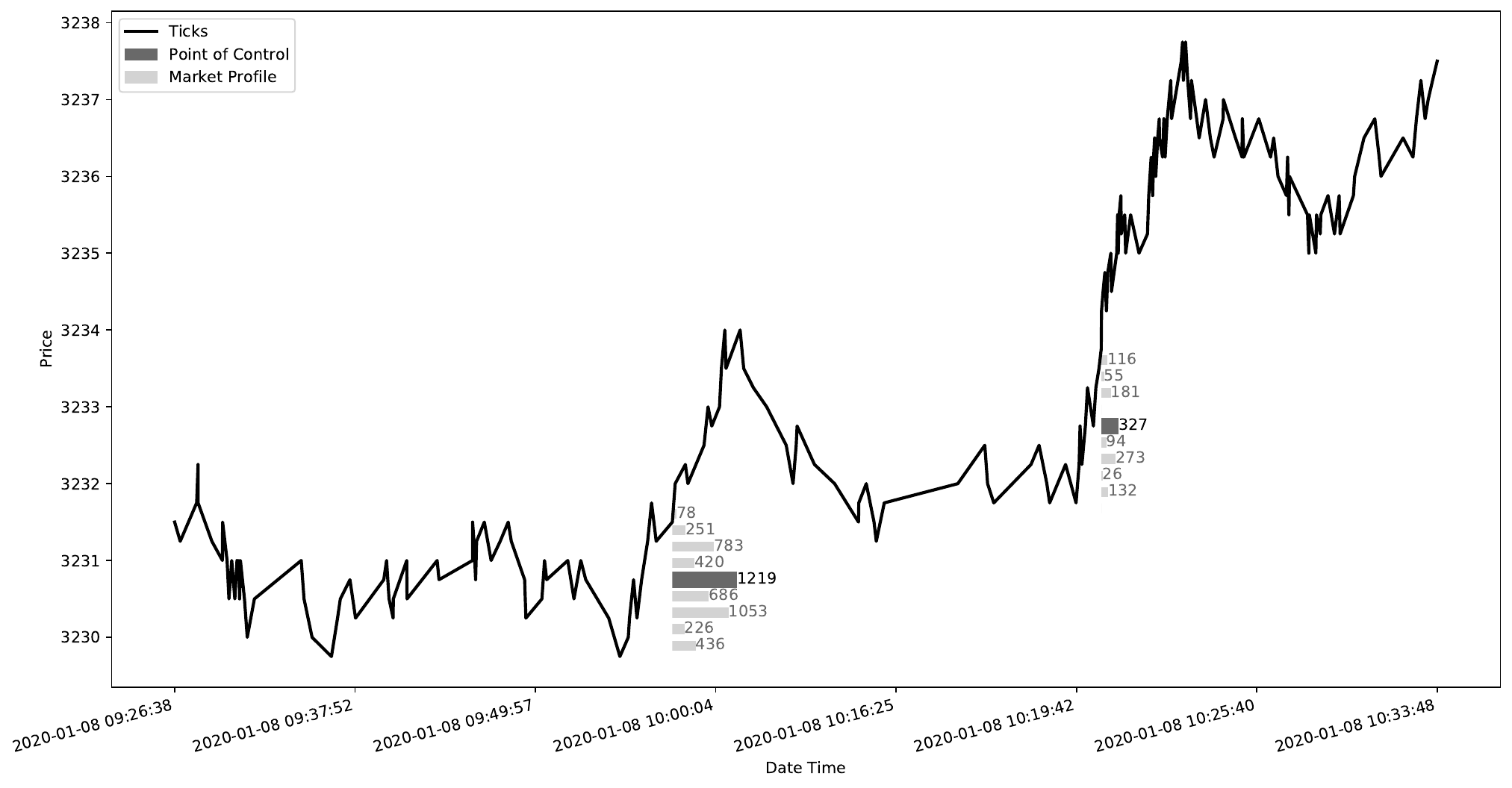}
    \caption{Example of volume-centred range bars generated for ES instrument. Histograms indicate traded volumes within the buffer. Points of control are in the centre of the volume profiles, marked with dark grey. The profiles are formed when the price buffer is complete (9 ticks in this case). Zero-volume entries are not shown.}
    \label{fig:VCRB}
\end{figure}

We generate VCRBs for range sizes of 5, 7, 9 and 11 ticks. For comparison purposes, we extract price-based patterns using a configuration suggested in \citet{sokolovsky2020machine}. 
We visualise the VCRBs in Fig~\ref{fig:VCRB}. As we stated in the methods, the volume in the centre is the largest one. The volume distributions differ a lot for the provided entries. If there is a price with zero volume (second volume profile in Figure~\ref{fig:VCRB}), we skip that price in the visualisation. 

\subsection{Prediction of the reversals and crossings}

The initial experimental stages are feature selection and model parameter tuning. We do not report the optimised feature spaces and model parameters in the manuscript, however, we make this data available in the shared reproducibility package~\citep{artur_sokolovsky_2021_4629568,sokolovsky_ncl_data}.

For the classification task, we report all the stated performance metrics for the configuration range 7, which is chosen based on RQ1 effect sizes, together with the PR-AUC metric for the price levels method in Tables~\ref{tab:perfMetricsES} and \ref{tab:perfMetricsB6}. The rest of the metrics for price levels are reported in the supplementary materials, in Tables~\ref{tab:SupplPLperfES} \& \ref{tab:SupplPLperfB6}. Additionally, in the supplementary materials we plot the data representing null and alternative hypotheses throughout the study - in Figures~\ref{fig:CBnoInfPrecisionSuppl}, \ref{fig:VCRBplPRAUCSuppl}, \ref{fig:VCRBesB6PRAUCSuppl} and~\ref{fig:FeatInteracBootstrapDistsSuppl}.

In all the experiment families we report effect sizes for all the VCRB configurations (ranges 5, 7, 9 and 11). The statistical test results are reported only for the largest effect size configuration from the RQ1 experiment family, on S\&P E-mini (ES) instrument. We report the rest of the statistical tests in the supplementary materials.
When evaluating the statistical tests, we correct for multiple comparisons - the corrected significance levels are provided separately for each experiment group.

\begin{table}[!htb]

\caption{Performance metrics for ES, volume-based pattern extraction configuration range 7. The dates are reported in the form MM/YY.}
    \centering
\begin{tabular}{l|ccccc|c}
{Batch} & \multicolumn{5}{c|}{\textbf{VCRB}} & \textbf{Price levels} \\
\hline
{} &  PR-AUC &  ROC-AUC &  F1-score &  Precision &  Null\_precision &  PR-AUC Price Levels \\
\hline
3/17 to 6/17  &    0.25 &     0.54 &      0.34 &       0.25 &            0.23 &                 0.14 \\
6/17 to 9/17  &    0.23 &     0.51 &      0.32 &       0.22 &            0.21 &                 0.19 \\
9/17 to 12/17 &    0.25 &     0.51 &      0.30 &       0.25 &            0.24 &                 0.15 \\
12/17 to 3/18 &    0.24 &     0.52 &      0.27 &       0.24 &            0.23 &                 0.16 \\
3/18 to 6/18  &    0.25 &     0.52 &      0.30 &       0.24 &            0.23 &                 0.16 \\
6/18 to 9/18  &    0.25 &     0.53 &      0.34 &       0.25 &            0.24 &                 0.16 \\
9/18 to 12/18 &    0.25 &     0.52 &      0.31 &       0.25 &            0.24 &                 0.16 \\
12/18 to 3/19 &    0.24 &     0.52 &      0.33 &       0.23 &            0.22 &                 0.18 \\
3/19 to 6/19  &    0.25 &     0.52 &      0.29 &       0.26 &            0.24 &                 0.16 \\
6/19 to 9/19  &    0.25 &     0.53 &      0.35 &       0.25 &            0.24 &                 0.15 \\
9/19 to 12/19 &    0.24 &     0.51 &      0.28 &       0.25 &            0.23 &                 0.14 \\
12/19 to 3/20 &    0.24 &     0.51 &      0.30 &       0.24 &            0.23 &                 0.17 \\
3/20 to 6/20  &    0.24 &     0.52 &      0.33 &       0.24 &            0.23 &                 0.17 \\
\end{tabular}

    \label{tab:perfMetricsES}
\end{table}

\begin{table}[!htb]
    \caption{Performance metrics for B6, volume-based pattern extraction configuration range 7. The dates are reported in the form MM/YY.}
    \centering
\begin{tabular}{l|ccccc|c}
& \multicolumn{5}{c|}{\textbf{VCRB}} & \textbf{Price levels} \\
\hline
{} &  PR-AUC &  ROC-AUC &  F1-score &  Precision &  Null\_precision &  PR-AUC \\
\hline
3/17 to 6/17  &    0.24 &     0.51 &      0.25 &       0.25 &            0.23 &                 0.17 \\
6/17 to 9/17  &    0.23 &     0.51 &      0.27 &       0.24 &            0.22 &                 0.16 \\
9/17 to 12/17 &    0.24 &     0.51 &      0.24 &       0.24 &            0.23 &                 0.16 \\
12/17 to 3/18 &    0.23 &     0.52 &      0.26 &       0.23 &            0.22 &                 0.15 \\
3/18 to 6/18  &    0.21 &     0.50 &      0.25 &       0.20 &            0.21 &                 0.12 \\
6/18 to 9/18  &    0.23 &     0.51 &      0.23 &       0.24 &            0.23 &                 0.18 \\
9/18 to 12/18 &    0.23 &     0.51 &      0.26 &       0.23 &            0.22 &                 0.15 \\
12/18 to 3/19 &    0.24 &     0.52 &      0.26 &       0.24 &            0.23 &                 0.19 \\
3/19 to 6/19  &    0.21 &     0.50 &      0.20 &       0.20 &            0.21 &                 0.18 \\
6/19 to 9/19  &    0.22 &     0.50 &      0.25 &       0.23 &            0.22 &                 0.20 \\
9/19 to 12/19 &    0.22 &     0.50 &      0.24 &       0.23 &            0.23 &                 0.14 \\
12/19 to 3/20 &    0.23 &     0.50 &      0.31 &       0.22 &            0.22 &                 0.17 \\
3/20 to 6/20  &    0.24 &     0.53 &      0.25 &       0.24 &            0.21 &                 0.20 \\
\end{tabular}

    \label{tab:perfMetricsB6}
\end{table}

\subsubsection{VCRB method, CatBoost versus No-information estimator (RQ1)}
\begin{figure}[!htb]
    \centering
    \includegraphics[width=0.75\textwidth]{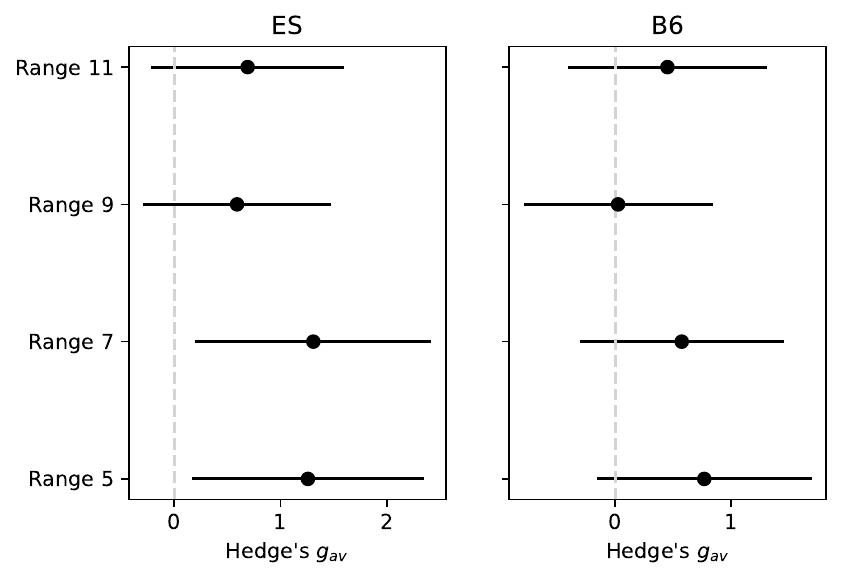}
    \caption{Hedge's g$_{av}$ effect sizes quantify the improvement of the  precision from using the CatBoost over the no-information estimator. The error bars illustrate the .95 confidence intervals, corrected for multiple comparisons. The dashed line corresponds to the significance threshold. Ranges correspond to different configurations of the pattern extraction method.}
    \label{fig:CBnoInfEffSize}
\end{figure}

Here we report the classification performance comparison between the CatBoost estimator and the no-information estimator. 
First, we plot the effect sizes with .95 confidence intervals in Fig~\ref{fig:CBnoInfEffSize}. Then, we provide performance statistics and the statistical test results in Table~\ref{tab:RQ1}. Additionally, to allow easier comprehension of the results, we visualise the precision of the models from Tables~\ref{tab:perfMetricsES} and~\ref{tab:perfMetricsB6} in supplementary materials, Fig~\ref{fig:CBnoInfPrecisionSuppl}.

\begin{table}[!htb]
    \caption{Statistics supporting the outcomes of the Wilcoxon test. The test is aimed to check whether, on the VCRB data and the considered feature space, CatBoost performs significantly better than the no-information estimator. The provided result is for the range 7 configuration.}
    \centering
    \begin{tabular}{c|c|c|c|c}
    \textbf{Statistics} & \multicolumn{4}{c}{\textbf{Dataset}} \\
    \hline
             & \multicolumn{2}{c}{\textbf{ES}} & \multicolumn{2}{|c}{\textbf{B6}} \\
             \hline
       One-tailed Wilcoxon test p-value & \multicolumn{2}{c}{$<.001$} & \multicolumn{2}{|c}{.024}\\
       Test Statistics & \multicolumn{2}{c}{91.0} & \multicolumn{2}{|c}{74.0}\\
       \hline
             & CatBoost & No-information & CatBoost & No-information \\
        
       Mean (precision) & 0.24 & 0.23 & 0.23 & 0.22 \\
       Median (precision) & 0.25 & 0.23 & 0.23 & 0.22 \\
       Standard Deviation (precision) & 0.0092 & 0.0074 & 0.0138 & 0.0067 \\
    \end{tabular}

    \label{tab:RQ1}
\end{table}

From Figure~\ref{fig:CBnoInfEffSize} we see that effect sizes for ES are generally larger. 
The effect size pattern across configurations is preserved between the two instruments with an exception of range 7. For ES the maximum effect size is observed at range 7 and minimum at range 9, while for B6 the maximum is at range 5 and the minimum at range 9.
Finally, judging the significance of the effect sizes by the confidence intervals (CIs) crossing the significance threshold line, we see that there are no significant effect sizes in B6, while ranges 5 and 7 are significant for ES. 

Performance statistics in Table~\ref{tab:RQ1} indicate no skew in the data between CatBoost and no-information models, at the same time there is a 1.2 and 2 times difference in the sample variance for ES and B6, respectively. The results for other VCRB configurations are provided in the supplementary materials, Table~\ref{tab:SupplRQ1}. 
Since 8 statistical tests were conducted (4 configurations $\times$ 2 instruments), we apply Bonferroni corrections to the significance level threshold for null hypothesis rejection. 
The corrected significance level is $\alpha=.05/8=.00625$.
For the rest of the experiment families, the statistical tests are reported for the range 7 configuration, as it demonstrated the largest effect size for RQ1 experiments. 

\subsubsection{VCRB versus price levels approach (RQ2)}
\begin{figure}[!htb]
    \centering
    \includegraphics[width=0.75\textwidth]{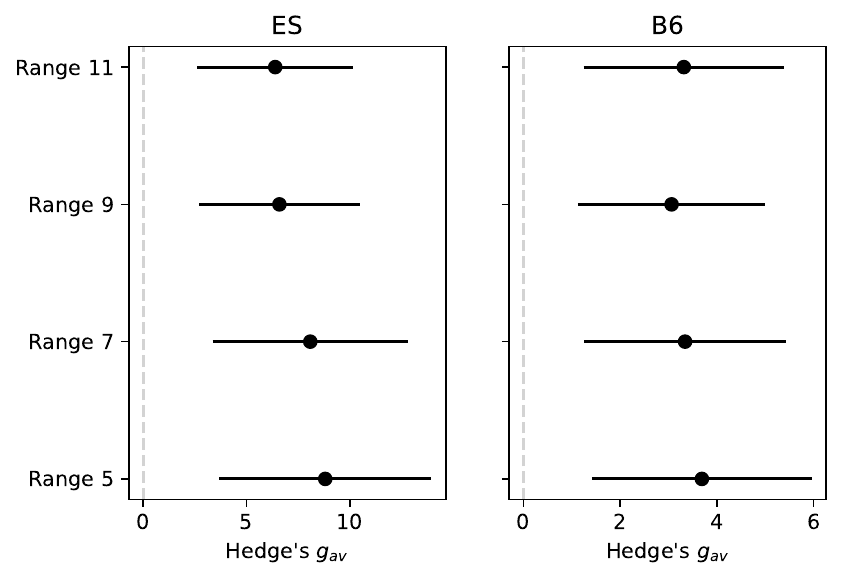}
    \caption{Hedge's \textit{g$_{av}$} effect sizes, quantifying the supremacy of the VCRB over the price levels approaches on the basis of the PR-AUC metric. The error bars illustrate the .95 confidence intervals, corrected for multiple comparisons. The dashed line accounts for the significance threshold. Ranges correspond to different configurations of the pattern extraction method.}
    \label{fig:VCRBplEffsize}
\end{figure}

Here the results of the investigation on whether volume-based centred bars lead to a better classification performance than the price level approach are presented. 
We report Hedge's \textit{g$_{av}$} effect sizes on paired data with .95 confidence intervals in Fig~\ref{fig:VCRBplEffsize}, and the outcomes of the statistical test with the supporting statistics in Table~\ref{tab:RQ2}. 
Complementing the results, we visualise PR-AUC values from Tables~\ref{tab:perfMetricsES} and~\ref{tab:perfMetricsB6} in supplementary materials, Fig~\ref{fig:VCRBplPRAUCSuppl}).

\begin{table}[!htb]
    \caption{Statistics support the outcomes of the Wilcoxon test which checks whether the Volume-based pattern extraction method leads to better classification performance than the price level pattern extraction. The result is reported for the range 7 configuration.}
    \centering
    \begin{tabular}{c|c|c|c|c}
    \textbf{Statistics} & \multicolumn{4}{c}{\textbf{Dataset}} \\
    \hline
             & \multicolumn{2}{c}{\textbf{ES}} & \multicolumn{2}{|c}{\textbf{B6}} \\
             \hline
       One-tailed Wilcoxon test p-value & \multicolumn{2}{c}{$<.001$} & \multicolumn{2}{|c}{$<.001$}\\
       Test Statistics & \multicolumn{2}{c}{91.0} & \multicolumn{2}{|c}{91.0}\\
       \hline
             & VCRB & Price levels & VCRB & Price levels \\
        
       Mean (PR-AUC) & 0.24 & 0.16 & 0.23 & 0.17 \\
       Median (PR-AUC) & 0.25 & 0.16 & 0.23 & 0.17 \\
       Standard Deviation (PR-AUC) & 0.0066 & 0.0118 & 0.0094 & 0.022 \\
    \end{tabular}

    \label{tab:RQ2}
\end{table}

One can see that all the effect sizes in Fig~\ref{fig:VCRBplEffsize} are significant as the confidence intervals do not overlap with the significance threshold line. Confidence intervals for ES are larger in comparison to B6. The largest effect size is observed for range 5 configuration across instruments. 

Performance statistics in Table~\ref{tab:RQ2} show no skew in the data, however, sample variances differ up to 4 times between the two methods. The rest of the VCRB configurations are reported in Table~\ref{tab:SupplRQ2}.
In the current experiment family, we run 8 tests in total, hence the corrected significance level is $\alpha=.05/8=.00625$. 

\subsubsection{VCRB method, ES versus B6 datasets (RQ3)}
\begin{figure}[!htb]
    \centering
    \includegraphics[width=0.65\textwidth]{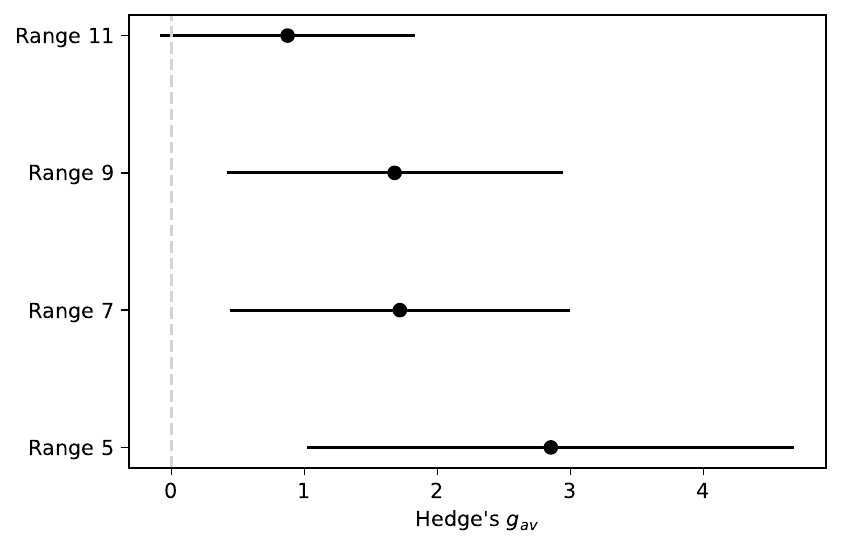}
    \caption{Hedge's g$_{av}$ effect sizes for Volume-based method PR-AUC performance improvement on ES over B6 datasets. Error bars illustrate the .95 confidence intervals corrected for multiple comparisons. Ranges correspond to different configurations of the pattern extraction method.}
    \label{fig:VCRBesB6Effsize}
\end{figure}

Here we detail the results of comparing the classification performance of the VCRB entries extracted from ES and B6 datasets. We report effect sizes with .95 confidence intervals in Fig~\ref{fig:VCRBesB6Effsize}, and the statistical test with the supporting statistics from range 7 configuration in Table~\ref{tab:RQ3}. The test outcomes for the other configurations are provided in supplementary materials, in Table~\ref{tab:SupplRQ3}. Additionally, we visualise PR-AUC performance from Tables~\ref{tab:perfMetricsES} and~\ref{tab:perfMetricsB6} in supplementary materials, Fig~\ref{fig:VCRBesB6PRAUCSuppl}.

\begin{table}[!htb]
    \caption{Statistics supporting the outcomes of the Wilcoxon test which assesses whether the VCRB pattern extraction method leads to better classification performance on the more liquid market (ES in comparison to B6). The result is reported for the range 7 configuration.}
    \centering
    \begin{tabular}{c|c|c}
    \textbf{Statistics} & \multicolumn{2}{c}{\textbf{Datasets}} \\
    \hline
       One-tailed Paired Wilcoxon test p-value & \multicolumn{2}{c}{$<.001$} \\
       Test Statistics & \multicolumn{2}{c}{88.0} \\
       \hline
             & ES & B6 \\
       Mean (PR-AUC) & 0.24 & 0.23 \\
       Median (PR-AUC) & 0.25 & 0.23 \\
       Standard Deviation (PR-AUC) & 0.0066 & 0.0094 \\
    \end{tabular}

    \label{tab:RQ3}
\end{table}

In Figure~\ref{fig:VCRBesB6Effsize} larger range of the VCRB leads to a smaller effect size. At the same time, the confidence intervals shrink with the range increase, indicating that the effect for the larger ranges is smaller but more stable. 

The statistics on the results in Table~\ref{tab:RQ3} show that variances differ by 50\% between the samples and there is no skew in the distributions. 
In the current experiment family, we run 4 tests in total, hence the corrected significance level is $\alpha=.05/4=.0125$. 

\subsection{Backtesting}
\begin{figure}[!htb]
    \centering
    \includegraphics[width=1.0\textwidth]{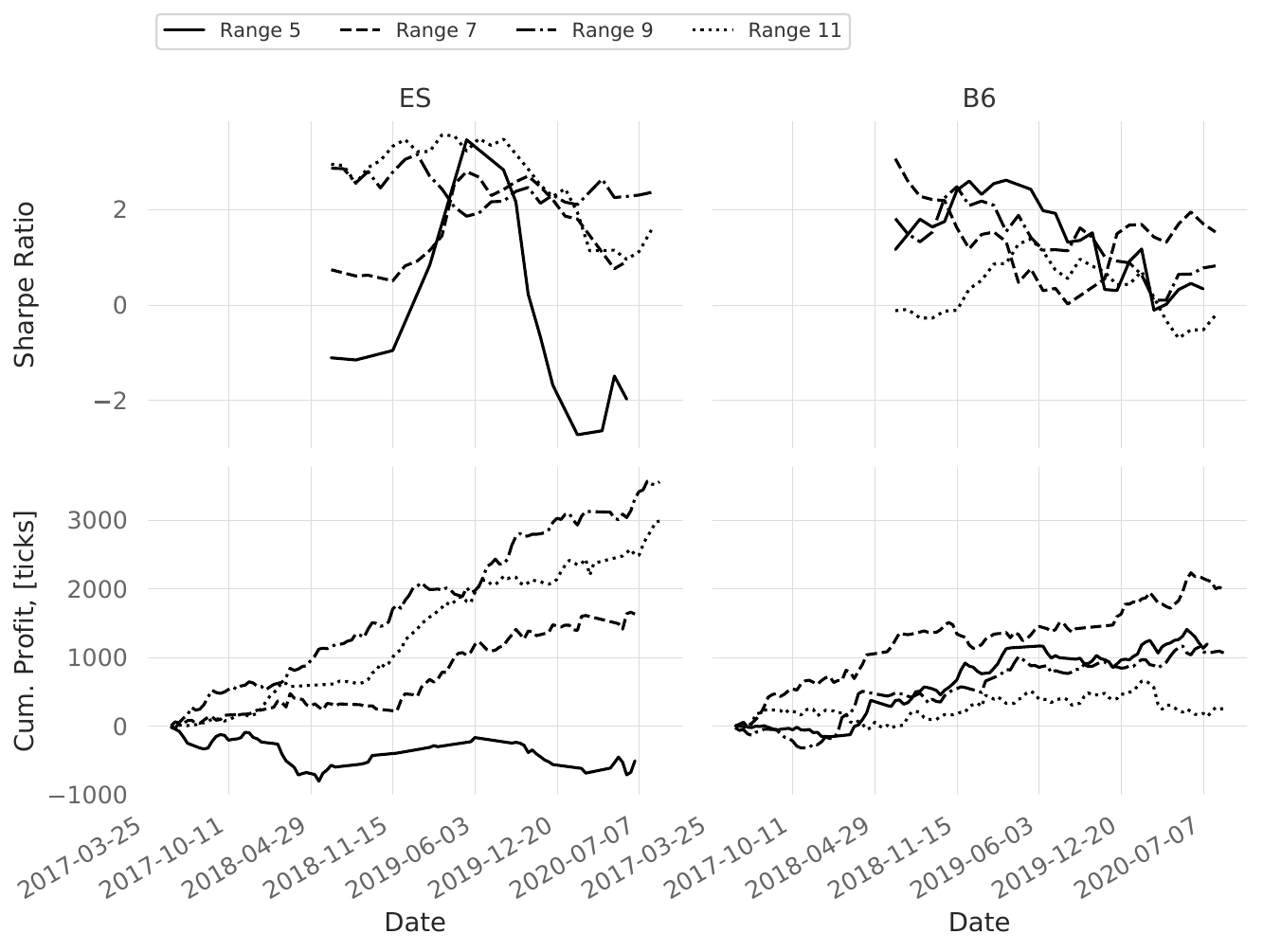}
    \caption{Sharpe Ratios and cumulative profits of the volume-based method configurations. Profits are provided in ticks. The simulation does not take bid-ask spreads and order queues into account, hence might be over-optimistic.}
    \label{fig:VOLsharpeProfit}
\end{figure}

\begin{figure}[!htb]
    \centering
    \includegraphics{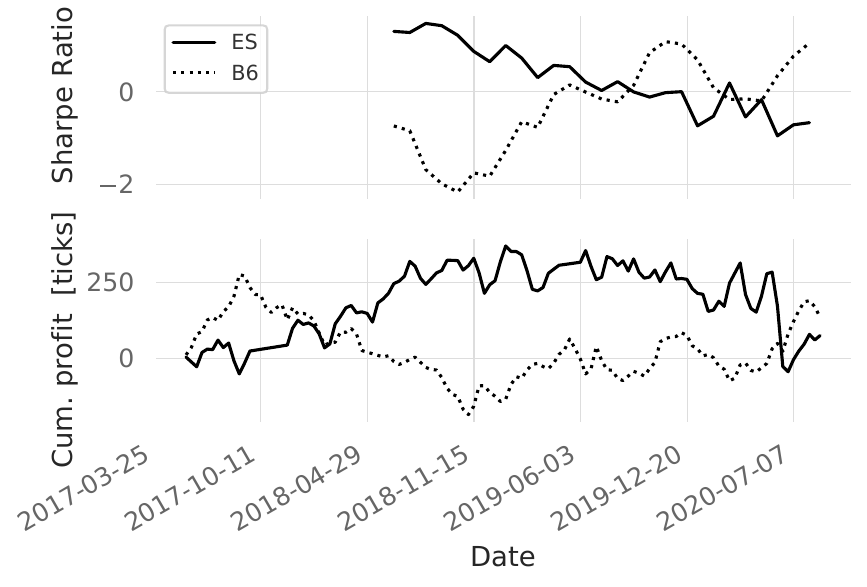}
    \caption{Sharpe ratios and cumulative profits of the price level-based method. Profits are provided in ticks. The simulation does not take bid-ask spreads and order queues into account, hence might be over-optimistic.}
    \label{fig:PLsharpeProfit}
\end{figure}

In Figures~\ref{fig:VOLsharpeProfit} and \ref{fig:PLsharpeProfit} we show annual rolling Sharpe Ratios with a 5\% risk-free rate and cumulative profits in ticks for all the configurations of the VCRB and price level methods.

For easier interpretation of the figures, we plot Sharpe Ratios averaged over 30-day periods. The lag of Sharpe Ratio plots with respect to the cumulative profits is caused by the requirement of the Sharpe Ratio metric to have a year of data available for obtaining the initial value.   

\subsection{Relatedness of feature interactions from SHAP and decision paths (RQ4)}
\begin{figure}[!htb]
    \centering
    \includegraphics[width=0.75\textwidth]{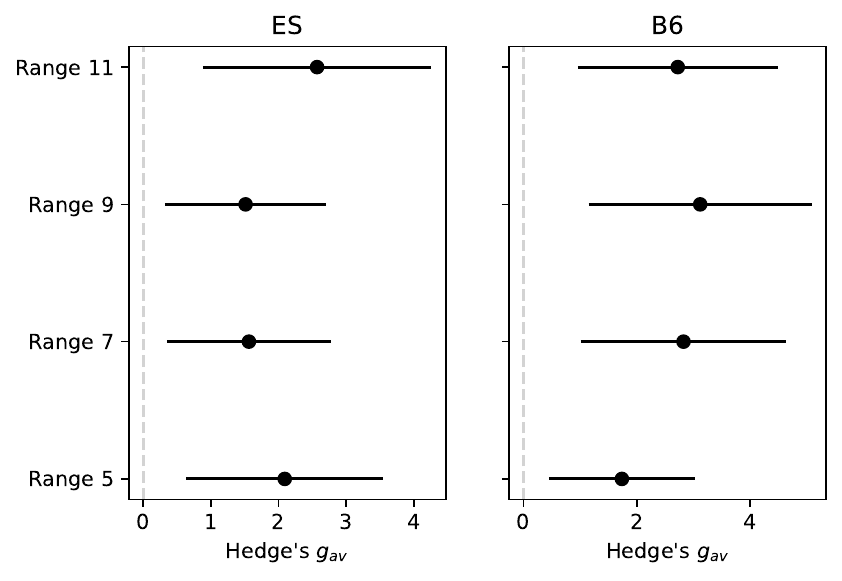}
    \caption{Hedge's g$_{av}$ effect sizes quantifying the relatedness strength of the SHAP and decision paths methods for extracting feature interactions with respect to the relatedness of the bootstrapped data. The relatedness of the feature interactions is assessed through the Footrule distances of the ranked interaction strengths. Error bars illustrate the .95 confidence intervals corrected for multiple comparisons. Ranges correspond to different configurations of the pattern extraction method.}
    \label{fig:SHAPmfEffsize}
\end{figure}

The following results assess the relatedness of the feature interactions data extracted using SHAP and the proposed decision paths methods by comparing their relatedness to the bootstrapped data. Following the format of the previous experiments, we report effect sizes in Figure~\ref{fig:SHAPmfEffsize}, and provide results of the statistical test as well as supporting statistics in Table~\ref{tab:RQ4}. Additionally, we report the statistical test outcomes for the rest of the configurations in the supplementary materials, Table~\ref{tab:SupplRQ4}. To allow easier assessment of the results, we plot the differences between data representing both hypotheses in supplementary materials, Figure~\ref{fig:FeatInteracBootstrapDistsSuppl}.

\begin{table}[!htb]
    \caption{Outcomes of the one-tailed Wilcoxon test which check whether SHAP and decision paths feature interactions extraction methods are related significantly stronger than the bootstrapped data. The statistics of the samples compared by the statistical test are provided in columns ''Actual distance'' \& ''Bootstrapped''. Mean, Median and Standard Deviation (SD) are produced for footrule distance. Footrule distance is inversely proportional to the relatedness. The result is reported for the range 7 configuration.}
    \centering
    \begin{tabular}{c|c|c|c|c}
    \textbf{Statistics} & \multicolumn{4}{c}{\textbf{Dataset}} \\
    \hline
             & \multicolumn{2}{c}{\textbf{ES}} & \multicolumn{2}{|c}{\textbf{B6}} \\
             \hline
       Wilcoxon & \multicolumn{2}{c}{$<.001$} & \multicolumn{2}{|c}{$<.001$}\\
       Test Statistics & \multicolumn{2}{c}{2.0} & \multicolumn{2}{|c}{0.0}\\
       \hline
             & Actual distance & Bootstrapped & Actual distance & Bootstrapped \\
        
       Mean & 88281 & 93281 & 86552 & 93280\\
       Median  & 88104 & 93279 & 86296 & 93281\\
       SD  & 4208.0 & 5.3 & 3134 & 6.6 \\
    \end{tabular}

    \label{tab:RQ4}
\end{table}

From Figure~\ref{fig:SHAPmfEffsize} we see that the smallest effect sizes are observed for configurations 7 and 9 in ES and for configuration 5 in B6. Interestingly, the behaviour across the configurations is flipped for ES and B6. 
In Table~\ref{tab:RQ4} one can see that there is no skew in the data as mean and median values are very similar. At the same time, we see significant differences in the data variances, which we address in the~\nameref{discussion} section. The test statistics values are small in comparison to the previous tests - in the current experiment smaller Footrule distances represent our alternative hypothesis, hence the statistical test is computed for the opposite difference sign with respect to the previous cases. 
In the current experiment family, we run 8 tests in total, hence the corrected significance level is $\alpha=.05/8=.00625$. 

\section{Discussion}
\label{discussion}
In the current section, we reflect on the obtained results in the same order as the experiments were conducted. 
Namely, a performance comparison between i) CatBoost and no-information models; ii) volume-based (VCRB) and price levels extraction methods; iii) ES and B6 financial instrument; iv) relatedness of feature interactions obtained from SHAP and the explicit model decision paths. 
Furthermore, we reflect on the limitations of the study as well as its broader implications and future work.

When discussing effect sizes, we note that in different fields interpretation of the effect sizes varies depending on the commonly observed differences in the effects between the test groups~\citep{Durlak2009HowSizes}. 
For instance, in social and medical sciences it is common to consider Hedge's \textit{g} effect sizes above 0.2, 0.5 and 0.8 as small, medium and large, respectively~\citep{Lakens2013CalculatingANOVAs}. 
To our knowledge, there are no established effect size thresholds in the field of financial time series data analysis. 
Hence, we mainly use the 0-threshold indicating absence of the effect size and contribute to the establishment of the domain-specific thresholds by reporting the effect sizes. Interpreting the data, we are guided by the confidence intervals as they represent a 95\% chance of finding the population effect size within the reported range. 
When considering the choice of effect size measure, it is important to distinguish between repeated measures and pair-matched comparisons. RQ1, RQ2 and RQ4 are good examples of repeated measures as we use the same data for extracting the patterns in two groups. RQ3 is an example of pair-matched comparison as we consider the same time intervals, and different but related time series for the two groups. Hence, we use Hedge's \textit{g$_{av}$} measure designed for paired data throughout the study. We note that using the original Hedge's \textit{g} measure in the RQ3 does not fundamentally change the results.  

Commenting on the potential profitability of the method, we compare the obtained Sharpe Ratios to the existing body of knowledge. For instance, a paper by Xiong et al.~\citep{xiong2018practical} reports the Sharpe Ratio of 1.79 using a deep reinforcement learning approach. The results of the study were obtained for US stocks.
Another study, by Yang et al.~\citep{yang2020deep}, reports the Sharpe Ratio of 1.3. In the study, the authors consider a subset of 30 US stocks data.
The approach suggested in the current study allows obtaining comparable performance, with the Sharpe Ratios being above 2 for certain experiment configurations. However, when directly comparing performance, the backtesting environment assumptions, as well as data granularity, should be taken into account. 
We highlight that the purpose of the current study is defined by the research questions, hence not aimed at maximisation of the strategy profitability. 

\subsection{RQ1 - Classification Performance of VCRB Bars}
Larger effect sizes for the ES instrument in Figure~\ref{fig:CBnoInfEffSize} mean that there is a larger improvement from using the CatBoost model for ES rather than for the B6 instrument. We reported the p-value of the statistical test in Tables~\ref{tab:RQ1} and~\ref{tab:SupplRQ1}. 
Considering the corrected significance level for the current experiment family, the null hypothesis is rejected for configurations 5 and 7, ES instrument (Tables~\ref{tab:RQ1} and \ref{tab:SupplRQ1}). The rejection of the null hypothesis means that the CatBoost estimator performs significantly better than the no-information model. 
From these results, we conclude that the feature space and the model work acceptably well in some configurations we can positively answer the first research question (RQ1) for the range 5 and 7 configurations. We note that the studied setting might have even more potential if considering extensive optimisation of the system parameters. 

No-information models, whose precision represents the fraction of the positively labelled patterns, are the highest for range 5 (Table~\ref{tab:SupplRQ1}). 
A potential interpretation is that this configuration is the most suitable for the studied markets. 
Possible underlying reasons include a lack of interest in the implied trading frequency from the larger market participants, whose capitals exceed the liquidity offered at this time scale. 
Alternatively, it might mean that the observed performance supremacy is purely theoretical and in reality, is levelled off by higher risks associated with more frequent market exposure (more trades).

Looking at the precision of the models in Figure~\ref{fig:CBnoInfPrecisionSuppl}, there is possibly a descending performance trend for both models and instruments. 
This can be interpreted as a gradual increase in market efficiency in the aspect of the wider use of market microstructure data for investment and trading decision making.
Overall, the CatBoost model with the considered feature space gives a larger performance increase with respect to the no-information model for ES than for the B6 dataset. 
This might be due to larger trading volumes of the ES instrument, hence less price action not supported by volumes. 

\subsection{RQ2 - Comparison of VCRB and Price Level Trading}
The statistical test outcomes are reported in Tables~\ref{tab:RQ2} and~\ref{tab:SupplRQ2}. 
Considering the corrected significance level for the current experiment family, all the statistical tests in the current experiment family have a significant outcome, and the null hypothesis is rejected. Answering the research question, the results mean that VCRB patterns can be classified with significantly better performance than price level-based patterns.

All the effect sizes in Figure~\ref{fig:VCRBplEffsize} are significant based on the confidence intervals not overlapping the 0-threshold. 
In Figure~\ref{fig:VCRBplPRAUCSuppl} we see quite a stable gap of around 0.08 in the PR-AUC performance between VCRB and price levels for ES. The gap is smaller (around 0.04) and converges for B6. 

Estimators might be worse at learning a reliable classification path from the price levels data for at least two hypothetical reasons: i) higher non-stationarity of the price level patterns in comparison to the VCRBs; ii) smaller price levels dataset sizes (Table~\ref{tab:extrData}). The latter can be potentially solved by increasing the considered time ranges of the training datasets. However, verification of any of these reasons is out of the scope of the current work and requires separate research for validation. 
Finally, we see a better backtesting performance for the volume-based method in comparison to the price level in Figs~\ref{fig:VOLsharpeProfit},~\ref{fig:PLsharpeProfit}, which additionally supports the findings.

\subsection{RQ3 - Impact of Market Liquidity on VCRB}
As per our initial hypothesis (H$_{13}$), VCRB trading performed significantly better in the more liquid market (ES).
This was shown by rejecting the null hypothesis, as the VCRB bars were able to classify with significantly better performance on the ES dataset for all the considered configurations (Tables~\ref{tab:RQ3} and~\ref{tab:SupplRQ3}).
Looking at the model performance for both instruments in Figure~\ref{fig:VCRBesB6PRAUCSuppl}, one sees that the method performs generally better for the ES dataset with up to 0.04 differences of PR-AUC, and two cases where B6 has a marginally better performance. 
This is something expected as with more liquidity comes more impact from the volume-based features.
This is also reflected in the number of Points of Control, with ever-increasing identified points correlating to an increase in liquidity (more recent years).

One of the potential reasons for the better performance of the ES dataset is a much larger number of extracted patterns in the ES market (Table~\ref{tab:extrData}), hence more training data is available.
Finally, we see a better backtesting performance for the more liquid asset in Fig~\ref{fig:VOLsharpeProfit}, which supports our formal findings.
Interestingly, we see that Sharpe Ratios have a similar character of changes across configurations, especially for B6. There is less similarity between the two pattern extraction methods. While it is a known fact that most financial instruments are related, making it hard to diversify risks, we see that these relations cause similar impacts on the trading performance. This might suggest that using the same trading approach across instruments might not contribute to risk diversification to the expected extent, and requires careful prior research. 

\subsection{RQ4 -  Feature Interaction Associations}
In this work, we investigated the relatedness of two different feature interaction methods.
SHAP values are a widely used measure that is easy to compute and approximates the predictor.
However, due to this approximation, our thinking was that it might be not suitable for every single setting, especially for the cases where the expected performance is far from ideal, like financial markets. 
Hence, we aimed at checking the relatedness of SHAP and explicitly extracted feature interactions. We constructed a null hypothesis dataset using a bootstrapping approach, if the relatedness between the null dataset and the other feature analysis dataset were similar then the relatedness would be insignificant; conversely, if the relatedness was akin between the interaction analysis methods but not the null hypothesis then the null hypothesis ($H_{04}$) would be rejected.
From the test outcomes in Tables~\ref{tab:RQ4} and \ref{tab:SupplRQ4} we conclude that for all the configurations and both instruments the null hypothesis is rejected and relatedness between the two methods is significant.
In Figure~\ref{fig:FeatInteracBootstrapDistsSuppl} we see that mean bootstrapped distances have a negligible variance in comparison to the actual feature interaction methods. 
The low variance of the null hypothesis data is a sign of the correct choice of the bootstrapped sample size. 
Also, the distances of all the entries are smaller than the null hypothesis data for B6 and ES with one exception entry being slightly larger than the bootstrapped entry.
We also have a strong relatedness between the two measures highlighting a strong interaction between the feature results of the two methods. Distances between the two methods are around 5\% smaller on average than the null hypothesis datasets. One should remember that the approaches have completely different underlying principles which probably causes these large differences.   
SHAP assigns a local explanation based on introducing a numeric measure of credit to each input feature. 
Then by combining many local explanations, it represents global structure while retaining local faithfulness to the original model~\citep{lundberg2020local}.
This is in contrast to our explicit approach which instead computes global values, accounting for the large mean distance.

\subsection{Limitations}
We would like to highlight that the analysis provided in this paper is but one of the possible means to empirically answer our research questions.
Whilst we choose to focus on strict statistical analysis to showcase the significance of the results, other valid approaches could be implemented.
A limitation of our work is that whilst our analysis is extensive, it focuses on only two examples of trading instruments.
Whilst these are chosen to represent different markets that can observe the performance of our method in vastly different scenarios, one could envision that a more thorough systematic analysis of the same approach across varied instruments could have some benefits; although we argue this may be out of the scope of the current work.
One potential limitation of our analysis resides in the relatively small sample size. To reject the null hypotheses we made use of the Wilcoxon test. 
This decision was guided by the small sample size, which in turn made it not feasible to reliably establish whether the data was normally distributed, consequently we could not use parametric tests such as t-tests.
Whilst sometimes less precise, this decision is supported by literature to be the optimum choice in these circumstances~\citep{skovlund2001should}.

When assessing the absolute model performance, we consider the theoretical profitability threshold computed for the simplistic strategy proposed in \citet{sokolovsky2020machine}. 
Concretely, assuming that the take-profit is 15 ticks, and the stop-loss is 3 ticks, we account 0.5 ticks for the trading fees (which is above a typical trading fee at the beginning of 2021). Here we do not take into account slippage as our approach uses limit orders for executing trades.
Hence, for each entry, we have a maximum possible theoretical profit of 14.5 ticks and a maximum loss of 3.5 ticks. Dividing one by another we get the theoretical profitability threshold at 24.1\% precision. Being more conservative, we would want to account for the bid-ask spread, which is 1 tick most of the time for ES and B6 (less stable). The presence of the spread means that after we enter the market, our open P\&L (profit \& loss) is -1 tick - if the position is opened by bid, it will be liquidated by ask which is 1 tick away and vice versa. 
Of course, it affects the fraction of the trades closed by the stop loss in a live setting in comparison to the no-spread simulation. We don't have enough information to probabilistically model this, but if we account for the spread by subtracting 1 tick from all trades, we end up with the profitability threshold of 33.3\% precision. While the original take profit to stop loss sizes relate as 1/5, the actual picture (after taking into account all the mechanics and fees) is very different.  
In our setting the limitation of the profitability threshold value comes from multiple entries observed within a short time range. With an already open position, the assumed strategy does not make use of the following signals until the current position gets liquidated. Also, there is a limitation caused by order queues hampering the strategy performance on the live market - potentially profitable orders are more likely to not be executed, as we are expecting the price to reverse. Losing positions will be executed always as the price continues its movement. We note that the different instruments and datasets will involve differences in volatility and liquidity, which impacts the length of the trades and other factors which may influence backtesting results. Considering all these, we conclude that it is necessary to take the backtesting results with a certain grain of salt.

We note that in some other machine learning contexts the obtained performance may seem low if not horrendous, however, in these very complex classification scenarios, it is in line with expectations as shown in previous works~\citep{sokolovsky2020machine,Dixon2017Classification-basedNetworks}. 

It should be noted that by design our hypotheses are tested on market microstructure-based feature space. By rejecting the null hypotheses we cannot claim that these findings hold for an arbitrary feature space, but rather for the proposed one. By running the experiments with the feature selection and model optimisation steps, we make an informal effort to expand the findings to a flexible feature set and a model configuration.
Also, even though we have made the best effort to unify the experiment design, the feature spaces slightly vary between the two methods, which may have some impact on the results.

Similar test statistics numbers across Tables~\ref{tab:RQ1}-\ref{tab:RQ3} are resulting from the small group sizes and similar pairwise measurement variable comparison outcomes. 

\subsection{Implications for practitioners}
The current study proposes an approach to the classification of patterns in multidimensional non-stationary financial time series. This work advances the body of knowledge in applied financial time series analysis by proposing a pattern extraction method and shaping the contexts in which it can be used. The proposed method can be directly applied as a part of an algorithmic trading pipeline. Moreover, the method might become an alternative way of sampling the market and stand in a row with other types of sampling, like volume and range bars.

There are other research areas dealing with a similar setting, like social networks and forums analysis, topic detection and tracking, fraud detection, etc. We believe that the idea of the proposed method is applicable to some of these fields. Namely, the identification of a pattern in one of the time series dimensions as an "anchor" for the multidimensional pattern extraction. For obtaining optimal results, the design of the anchor should involve domain knowledge. 

\subsection{Future work}

We see a number of paths for future work. Namely, an extension of the experiments to other markets, more advanced backtesting, fundamental assessment of the stationarity, and extension of the feature space. 
Other financial markets, like Forex, Crypto and stocks would require certain adjustments to the proposed method since there are multiple marketplaces and volumes are distributed across them. Moreover, prices might also differ between the marketplaces, making it harder to aggregate the data. Overall, each financial instrument has its own characteristic properties and it would be interesting to see how generalisable the proposed method is.

While the used backtesting engine is relatively simple, there are more advanced ones exist in the field. However, they are usually made available as parts of trading platforms. Since the trading platforms are usually proprietary, the mechanics of the backtesting engines are not always clear and transparent and cannot be replicated outside the platform. Hence, it would be beneficial to develop an open-source package for backtesting which includes bid-ask spreads and models of the trading queues. 

Our study is based on empirical methods and is considered application-oriented. We hypothesise that the proposed method allows extracting patterns that are more stationary than the market itself. However, we never formally measure the stationarity, to avoid further complications of the study design. A formal assessment of the stationarity would allow choosing the most promising pattern extraction methods which in theory might require fewer training entries for successful classification and trading. 

In the current study, we used volumes-based feature space. There is a different approach to feature design - technical indicators, like RSI, MACD, Parabolic SAR, etc. To fully incorporate the indicators into the pipeline, the feature space should be increased significantly. This might be done in parallel with the more fine-grained configuration of the estimator and the development of a more complex trading strategy. Even though this point is rather implementation-focused, it might lead to very interesting results, which could be further supported by the statistical approach followed in the current study. 

\section{Conclusion}
\label{sec:conclusion}
In this study, we present a new market pattern extraction method suitable for ML called Volume-Centred Range Bars (VCRB).
The study presents a detailed statistical analysis of the presented approach to thoroughly assess its performance.

We firstly assess the volume-based pattern extraction validity by evaluating 1) the significance of the classification performance, 2) the improvement of the proposed feature space and 3) model configurations with respect to the baseline (RQ1).
This expands beyond simply trading using VCRBs as we showcase how performance can be improved using a state-of-the-art feature engineering approach and a machine learning estimator.
We further investigate the method's effectiveness by comparing it with another successful pattern extraction method based on price levels.
The results showcase a net improvement in performance across two different financial instruments (RQ2).
By rejecting the null hypothesis H$_{03}$, we answer positively in our research question 3 (RQ3) that liquid markets improve the effectiveness of the proposed approach. 

Additionally, contributing to the explainability, we compare two different feature interaction extraction approaches - the popular ML approach SHAP, which approximates the model, and an extension of Monoforest, which uses explicit decision paths from the model.
The analysis shows that in the considered setting, both methods are significantly related, hence holding some common findings. We conclude that SHAP is effective in providing explainability in the considered setting, something which had previously not been investigated.

To conclude, our methodology is structured in a way that allows for comparability across studies by providing the effect sizes; and reproducibility, by detailing the method and sharing the reproducibility package. 
Our hope is that this will make it easier for the practitioners to test this same approach and evaluate it against other methods, hopefully helping improve the field for the better.
Code to reproduce our analysis and match our results is available online~\citep{artur_sokolovsky_2021_4629568,sokolovsky_ncl_data}.

\section*{Acknowledgement}
The research is supported by CRITiCaL - Combatting cRiminals In The CLoud, under grant EP/M020576/1, and by the AISEC grant under EPSRC number EP/T027037/1

\bibliography{mybibfile.bib, references.bib}

\section*{Supplementary Materials}
\label{tab:glossary}
\setcounter{table}{0}
\renewcommand\thetable{S\arabic{table}}
\setcounter{figure}{0}
\renewcommand\thefigure{S\arabic{figure}}

\section*{Glossary}
       \begin{longtable}{p{0.2\linewidth} | p{0.8\linewidth}}
        \hline
        \textbf{Terms} & \textbf{Definitions} \\
        \cline{0-1}
        \textit{Futures contract} & provide means to trade a commodity (instrument) at a predetermined price at a specific time in the future. \\ \hline
        \textit{Tick} &  represent a single movement upward or downward by a specific increment in the price for a specific instrument (e.g. 0.25\$ for S\&P futures).\\ \hline
        \textit{Bar} & used to identify a window of interest based on some heuristic, and then aggregate the features of that window. 
        It May contain several features and it is up to the individual to decide what features to select, common features include: \textit{Bar start time}, \textit{Bar end time}, \textit{Sum of Volume}, \textit{Open Price}, \textit{Close Price}, \textit{Min} and \textit{Max} (usually called High and Low) prices, and any other features that might help characterise the trading performed within this window\\\hline
        \textit{Volume} & refers to the number of traded contracts (or shares) for a particular instrument\\ \hline
        \textit{Volume profile} & refers to the volumes traded per price visualised as a vertical histogram for a range of prices over a certain time range\\ \hline
        \textit{Liquidity} & how rapidly stocks may be traded without affecting the market price. Has an impact on whether you are able to get the desired instrument at your choice of price (sell or buy)\\ \hline
        \textit{Volatility} & degree of variation for the price of a given instrument over a period of time. \\ \hline
        \textit{Trading} &  the buying and selling of an instrument\\ \hline
        \textit{A trading platform} & is software that you use to conduct your trading. Allows for the centralised management of instruments and positions\\\hline
        \textit{Time \& Sales} &  a set of features provided real-time for each trade executed in an exchange. Features include: \textit{volume}, \textit{price}, \textit{direction}, \textit{date}, and \textit{time} \\\hline
        \textit{Order Book} & the list of orders used by a trading venue to keep track of offers and bids by buyers and sellers for a particular instrument. These are then matched in specific order to execute a trade \\\hline
        \textit{Flat Market} & is a stable state in which the range for the broader market does not move either higher or lower, but instead trades within the boundaries of recent highs and lows.\\\hline 
        \textit{Trending Market} & shifts in the market towards a raise or decrease in price compared to expected highs or lows. Used to buy and sell at the point where ones are most likely to gain profit\\\hline
        \textit{Long positions} & owning the asset for a time period with the expectation that the asset will go up in price\\\hline
        \textit{Short positions} &  if the expectation is that the price will decrease over time, you can short an asset to profit from its decreasing value.\\\hline
        \textit{Actionable ML} & In the context of machine learning and more specifically algorithmic trading, actionability refers to the ability to act upon a prediction. This may directly relate to understanding the reason behind the prediction, in turn, allowing you to make informed decisions on how to act upon it.  \\ \hline
        \textit{Take-profit} & is an order that specifies a price at which to liquidate a profitable trade. The order remains open until the price is reached\\ \hline
        \textit{Stop-loss} & is an order placed at a specific price that gets executed if the position losses reach a certain threshold. This is meant to explicitly manage the allowed losses per trade. \\ 
        \hline
        \caption*{\textbf{Glossary}: This glossary contains some essential definitions used throughout the paper. Sometimes similar definitions are reintroduced in specific contexts to centre the discussion.}
    \label{tab:glossary}
\end{longtable}

\begin{table}[]
    \caption{Performance metrics for ES, price levels pattern extraction method. The dates are reported in the form MM/YY. "Null\_precision" corresponds to the no-information model precision.}
    \centering
\begin{tabular}{l|ccccc}
{} &  PR-AUC &  ROC-AUC &  F1-score &  Precision &  Null\_precision \\
\hline
3/17 to 6/17  &    0.14 &     0.44 &      0.11 &       0.10 &            0.15 \\
6/17 to 9/17  &    0.19 &     0.53 &      0.14 &       0.14 &            0.18 \\
9/17 to 12/17 &    0.15 &     0.49 &      0.20 &       0.16 &            0.16 \\
12/17 to 3/18 &    0.16 &     0.50 &      0.17 &       0.15 &            0.16 \\
3/18 to 6/18  &    0.16 &     0.48 &      0.07 &       0.12 &            0.17 \\
6/18 to 9/18  &    0.16 &     0.51 &      0.25 &       0.17 &            0.16 \\
9/18 to 12/18 &    0.16 &     0.51 &      0.10 &       0.16 &            0.15 \\
12/18 to 3/19 &    0.18 &     0.53 &      0.13 &       0.18 &            0.17 \\
3/19 to 6/19  &    0.16 &     0.50 &      0.18 &       0.17 &            0.17 \\
6/19 to 9/19  &    0.15 &     0.50 &      0.14 &       0.14 &            0.15 \\
9/19 to 12/19 &    0.14 &     0.50 &      0.14 &       0.13 &            0.15 \\
12/19 to 3/20 &    0.17 &     0.50 &      0.15 &       0.18 &            0.17 \\
3/20 to 6/20  &    0.17 &     0.51 &      0.21 &       0.17 &            0.16 \\
\end{tabular}

    \label{tab:SupplPLperfES}
\end{table}

\begin{table}[]
    \caption{Performance metrics for B6, price levels pattern extraction method. "Null\_precision" corresponds to the no-information model precision.}
    \centering
\begin{tabular}{l|ccccc}
{} &  PR-AUC &  ROC-AUC &  F1-score &  Precision &  Null\_precision \\
\hline
3/17 to 6/17  &    0.17 &     0.47 &      0.15 &       0.15 &            0.17 \\
6/17 to 9/17  &    0.16 &     0.50 &      0.15 &       0.18 &            0.16 \\
9/17 to 12/17 &    0.16 &     0.49 &      0.24 &       0.16 &            0.16 \\
12/17 to 3/18 &    0.15 &     0.48 &      0.15 &       0.17 &            0.16 \\
3/18 to 6/18  &    0.12 &     0.51 &      0.13 &       0.11 &            0.13 \\
6/18 to 9/18  &    0.18 &     0.54 &      0.27 &       0.20 &            0.16 \\
9/18 to 12/18 &    0.15 &     0.50 &      0.18 &       0.13 &            0.13 \\
12/18 to 3/19 &    0.19 &     0.51 &      0.22 &       0.14 &            0.15 \\
3/19 to 6/19  &    0.18 &     0.53 &      0.11 &       0.18 &            0.16 \\
6/19 to 9/19  &    0.20 &     0.51 &      0.21 &       0.22 &            0.19 \\
9/19 to 12/19 &    0.14 &     0.45 &      0.13 &       0.15 &            0.16 \\
12/19 to 3/20 &    0.17 &     0.50 &      0.20 &       0.22 &            0.15 \\
3/20 to 6/20  &    0.20 &     0.52 &      0.12 &       0.14 &            0.16 \\
\end{tabular}

    \label{tab:SupplPLperfB6}
\end{table}
\begin{table}[!htb]
    \caption{Statistics supporting the outcomes of the Wilcoxon test. The test is aimed to validate whether on the VCRB data and the considered feature space, CatBoost performs significantly better than the no-information estimator. The result is reported for the range configurations of 5, 9 and 11.}
    \centering
    \begin{tabular}{c|c|c|c|c}
    {Statistics} & \multicolumn{4}{c}{{Dataset}} \\
    \hline
    & \multicolumn{4}{c}{\textbf{Range 5}} \\
             & \multicolumn{2}{c}{{ES}} & \multicolumn{2}{|c}{{B6}} \\
             \hline
       One-tailed Wilcoxon test p-value & \multicolumn{2}{c}{$<.001$} & \multicolumn{2}{|c}{.024}\\
       Test Statistics & \multicolumn{2}{c}{90.0} & \multicolumn{2}{|c}{74.0}\\
       \hline
             & CatBoost & No-information & CatBoost & No-information \\
        
       Mean (precision) & 0.25 & 0.24 & 0.23 & 0.23 \\
       Median (precision) & 0.25 & 0.24 & 0.23 & 0.23 \\
       Standard Deviation (precision) & 0.0084 & 0.0049 & 0.0081 & 0.0039 \\
       \hline
    \end{tabular}
    
    \begin{tabular}{c|c|c|c|c}
    \hline
    & \multicolumn{4}{c}{\textbf{Range 9}} \\
             & \multicolumn{2}{c}{{ES}} & \multicolumn{2}{|c}{{B6}} \\
             \hline
       One-tailed Wilcoxon test p-value & \multicolumn{2}{c}{.0199} & \multicolumn{2}{|c}{.47}\\
       Test Statistics & \multicolumn{2}{c}{75.0} & \multicolumn{2}{|c}{47.0}\\
       \hline
             & CatBoost & No-information & CatBoost & No-information \\
        
       Mean (precision) & 0.24 & 0.23 & 0.22 & 0.22 \\
       Median (precision) & 0.24 & 0.23 & 0.23 & 0.22 \\
       Standard Deviation (precision) & 0.0109 & 0.0043 & 0.0112 & 0.0068 \\
       \hline
    \end{tabular}
    
\begin{tabular}{c|c|c|c|c}

    \hline
    & \multicolumn{4}{c}{\textbf{Range 11}} \\
             & \multicolumn{2}{c}{{ES}} & \multicolumn{2}{|c}{{B6}} \\
             \hline
       One-tailed Wilcoxon test p-value & \multicolumn{2}{c}{.0164} & \multicolumn{2}{|c}{.170}\\
       Test Statistics & \multicolumn{2}{c}{76.0} & \multicolumn{2}{|c}{60.0}\\
       \hline
             & CatBoost & No-information & CatBoost & No-information \\
        
       Mean (precision) & 0.24 & 0.24 & 0.23 & 0.23 \\
       Median (precision) & 0.24 & 0.23 & 0.23 & 0.23 \\
       Standard Deviation (precision) & 0.0118 & 0.0057 & 0.0201 & 0.0082 \\
    \end{tabular}

    \label{tab:SupplRQ1}
\end{table}
\begin{table}[!htb]
    \caption{Statistics supporting the outcomes of the Wilcoxon test which validates whether Volume-based pattern extraction method leads to better classification performance than the price level pattern extraction. The result is reported for the range configurations of 5, 9 and 11.}
    \centering
\begin{tabular}{c|c|c|c|c}
    {Statistics} & \multicolumn{4}{c}{{Dataset}} \\
    \hline
    & \multicolumn{4}{c}{\textbf{Range 5}} \\
             & \multicolumn{2}{c}{{ES}} & \multicolumn{2}{|c}{{B6}} \\
             \hline
       One-tailed Wilcoxon test p-value & \multicolumn{2}{c}{$<.001$} & \multicolumn{2}{|c}{$<.001$}\\
       Test Statistics & \multicolumn{2}{c}{91.0} & \multicolumn{2}{|c}{91.0}\\
       \hline
                          & VCRB & Price levels & VCRB & Price levels \\
        
       Mean (PR-AUC) & 0.25 & 0.16 & 0.23 & 0.17 \\
       Median (PR-AUC) & 0.25 & 0.16 & 0.23 & 0.17 \\
       Standard Deviation (PR-AUC) & 0.0066 & 0.0118 & 0.0060 & 0.022 \\
       \hline
    \end{tabular}
    
    \begin{tabular}{c|c|c|c|c}
    \hline
    & \multicolumn{4}{c}{\textbf{Range 9}} \\
             & \multicolumn{2}{c}{{ES}} & \multicolumn{2}{|c}{{B6}} \\
             \hline
       One-tailed Wilcoxon test p-value & \multicolumn{2}{c}{$<.001$} & \multicolumn{2}{|c}{$<.001$}\\
       Test Statistics & \multicolumn{2}{c}{91.0} & \multicolumn{2}{|c}{91.0}\\
       \hline
                          & VCRB & Price levels & VCRB & Price levels \\
        
       Mean (PR-AUC) & 0.24 & 0.16 & 0.22 & 0.17 \\
       Median (PR-AUC) & 0.24 & 0.16 & 0.22 & 0.17 \\
       Standard Deviation (PR-AUC) & 0.0106 & 0.0118 & 0.0085 & 0.022 \\
       \hline
    \end{tabular}
    
\begin{tabular}{c|c|c|c|c}
    \hline
    & \multicolumn{4}{c}{\textbf{Range 11}} \\
             & \multicolumn{2}{c}{{ES}} & \multicolumn{2}{|c}{{B6}} \\
             \hline
       One-tailed Wilcoxon test p-value & \multicolumn{2}{c}{$<.001$} & \multicolumn{2}{|c}{$<.001$}\\
       Test Statistics & \multicolumn{2}{c}{91.0} & \multicolumn{2}{|c}{91.0}\\
       \hline
                          & VCRB & Price levels & VCRB & Price levels \\
        
       Mean (PR-AUC) & 0.24 & 0.16 & 0.23 & 0.17 \\
       Median (PR-AUC) & 0.24 & 0.16 & 0.23 & 0.17 \\
       Standard Deviation (PR-AUC) & 0.0126 & 0.0118 & 0.0131 & 0.022 \\
    \end{tabular}

    \label{tab:SupplRQ2}
\end{table}
\begin{table}[!htb]
    \caption{Statistics supporting the outcomes of the Wilcoxon test which assesses whether VCRB pattern extraction method leads to better classification performance on the more liquid market (ES in comparison to B6). The result is reported for the range configurations of 5, 9 and 11.}
    \centering
    \begin{tabular}{c|c|c}
    {Statistics} & \multicolumn{2}{c}{Datasets} \\
    \hline
   & \multicolumn{2}{c}{\textbf{Range 5}} \\
       One-tailed Wilcoxon test p-value & \multicolumn{2}{c}{$<.001$} \\
       Test Statistics & \multicolumn{2}{c}{91.0} \\
       \hline
             & ES & B6 \\
       Mean (PR-AUC) & 0.25 & 0.23 \\
       Median (PR-AUC) & 0.25 & 0.23 \\
       Standard Deviation (PR-AUC) & 0.0066 & 0.0060 \\
       \hline
    \end{tabular}
    
    \begin{tabular}{c|c|c}
    \hline
    & \multicolumn{2}{c}{\textbf{Range 9}} \\
       One-tailed Wilcoxon test p-value & \multicolumn{2}{c}{$<.001$} \\
       Test Statistics & \multicolumn{2}{c}{88.0} \\
       \hline
             & ES & B6 \\
       Mean (PR-AUC) & 0.24 & 0.22 \\
       Median (PR-AUC) & 0.24 & 0.22 \\
       Standard Deviation (PR-AUC) & 0.0106 & 0.0085 \\
       \hline
    \end{tabular}
    
    \begin{tabular}{c|c|c}
    \hline
    & \multicolumn{2}{c}{\textbf{Range 11}} \\
       One-tailed Wilcoxon test p-value & \multicolumn{2}{c}{.040} \\
       Test Statistics & \multicolumn{2}{c}{71.0} \\
       \hline
             & ES & B6 \\
       Mean (PR-AUC) & 0.24 & 0.23 \\
       Median (PR-AUC) & 0.24 & 0.23 \\
       Standard Deviation (PR-AUC) & 0.0126 & 0.0131 \\
    \end{tabular}

    \label{tab:SupplRQ3}
\end{table}
\begin{table}[!htb]
    \caption{Statistics supporting the outcomes of the Wilcoxon test which assesses whether SHAP and decision paths feature interaction extraction methods are related significantly stronger than the bootstrapped data. Footstep distance is inversely proportional to the relatedness. The result is reported for the range configurations of 5, 9 and 11.}
    \centering
    \begin{tabular}{c|c|c|c|c}
    {Statistics} & \multicolumn{4}{c}{{Dataset}} \\
    \hline
    & \multicolumn{4}{c}{\textbf{Range 5}} \\
             & \multicolumn{2}{c}{{ES}} & \multicolumn{2}{|c}{{B6}} \\
             \hline
       One-tailed Wilcoxon test p-value & \multicolumn{2}{c}{$<.001$} & \multicolumn{2}{|c}{$<.001$}\\
       Test Statistics & \multicolumn{2}{c}{3.0} & \multicolumn{2}{|c}{2.0}\\
       \hline
             & Actual distance & Bootstrapped & Actual distance & Bootstrapped \\
        
       Mean (footstep distance) & 88247 & 93280 & 87891 & 93279\\
       Median (footstep distance) & 87390 & 93281 & 88408 & 93278\\
       Standard Deviation (footstep distance) & 3166.0 & 5.0 & 4083.0 & 3.4 \\
       \hline
    \end{tabular}
    
    \begin{tabular}{c|c|c|c|c}
    & \multicolumn{4}{c}{\textbf{Range 9}} \\
             & \multicolumn{2}{c}{{ES}} & \multicolumn{2}{|c}{{B6}} \\
             \hline
       One-tailed Wilcoxon test p-value & \multicolumn{2}{c}{.00171} & \multicolumn{2}{|c}{$<.001$}\\
       Test Statistics & \multicolumn{2}{c}{6.0} & \multicolumn{2}{|c}{0.0}\\
       \hline
             & Actual distance & Bootstrapped & Actual distance & Bootstrapped \\
        
       Mean (footstep distance) & 88188 & 93279 & 86172 & 93278\\
       Median (footstep distance) & 88728 & 93279 & 86132 & 93279\\
       Standard Deviation (footstep distance) & 4429.0 & 6.1 & 2998.0 & 4.6\\
       \hline
    \end{tabular}
    
    \begin{tabular}{c|c|c|c|c}
    & \multicolumn{4}{c}{\textbf{Range 11}} \\
             & \multicolumn{2}{c}{{ES}} & \multicolumn{2}{|c}{{B6}} \\
             \hline
       One-tailed Wilcoxon test p-value & \multicolumn{2}{c}{$<.001$} & \multicolumn{2}{|c}{$<.001$}\\
       Test Statistics & \multicolumn{2}{c}{1.0} & \multicolumn{2}{|c}{0.0}\\
       \hline
             & Actual distance & Bootstrapped & Actual distance & Bootstrapped \\
        
       Mean (footstep distance) & 87266 & 93281 & 87228 & 93280\\
       Median (footstep distance) & 86994 & 93281 & 86198 & 93279\\
       Standard Deviation (footstep distance) & 3081.0 & 5.5 & 2925.0 & 3.4 \\
    \end{tabular}

    \label{tab:SupplRQ4}
\end{table}

\begin{figure}[!htb]
    \centering
    \includegraphics[width=\textwidth]{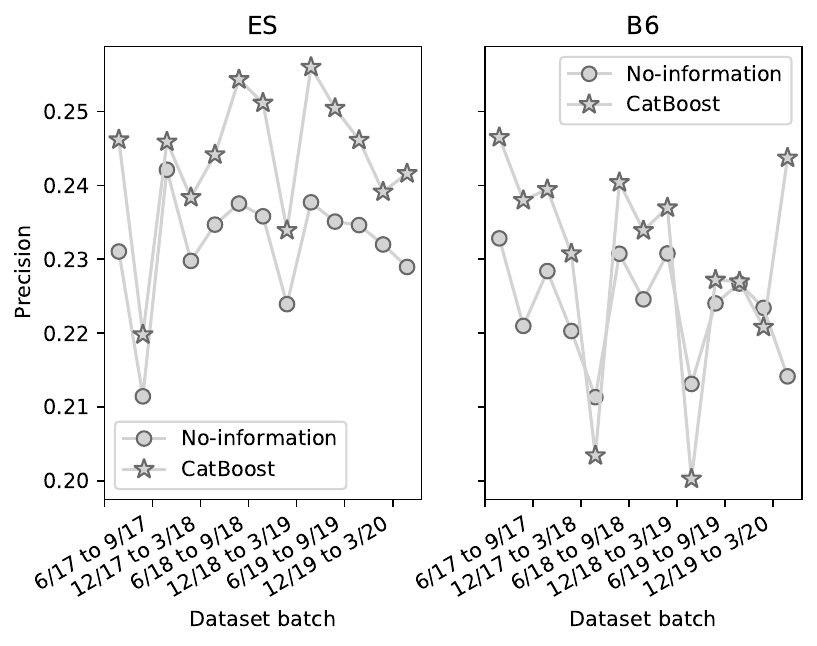}
    \caption{Precision performance metric for the no-information model and CatBoost plotted for both instruments - ES and B6. The Y-axis is mutual for both subplots.}
    \label{fig:CBnoInfPrecisionSuppl}
\end{figure}

\begin{figure}[!htb]
    \centering
    \includegraphics[width=\textwidth]{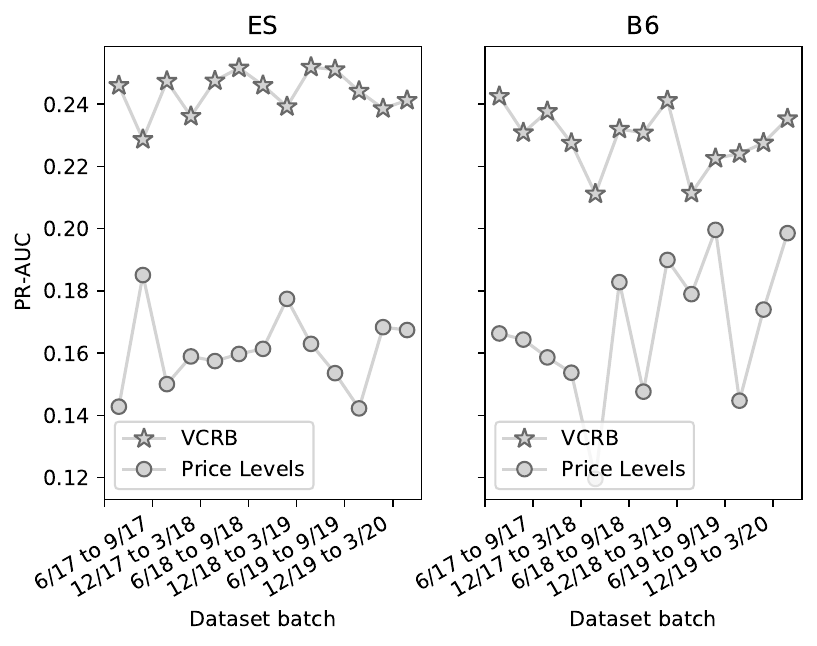}
    \caption{We plot PR-AUC for volume-based pattern extraction method and price level-based. The metric is reported for ES and B6 instruments. Volume-based method is reported for range 7 configuration.}
    \label{fig:VCRBplPRAUCSuppl}
\end{figure}

\begin{figure}[!htb]
    \centering
    \includegraphics[width=\textwidth]{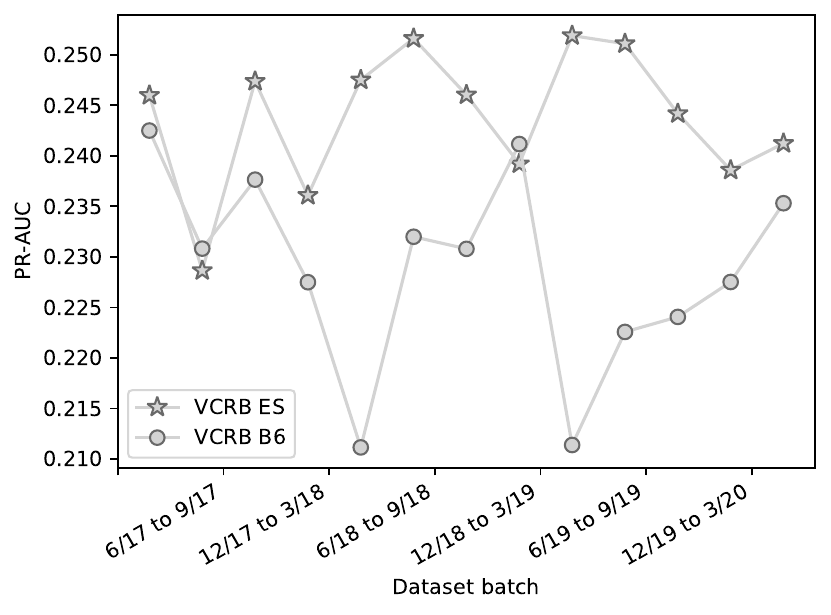}
        \caption{PR-AUC of CatBoost models obtained for ES and B6 futures instruments. Reported for range 7 configuration.}
    \label{fig:VCRBesB6PRAUCSuppl}
\end{figure}

\begin{figure}[!htb]
    \centering
    \includegraphics[width=\textwidth]{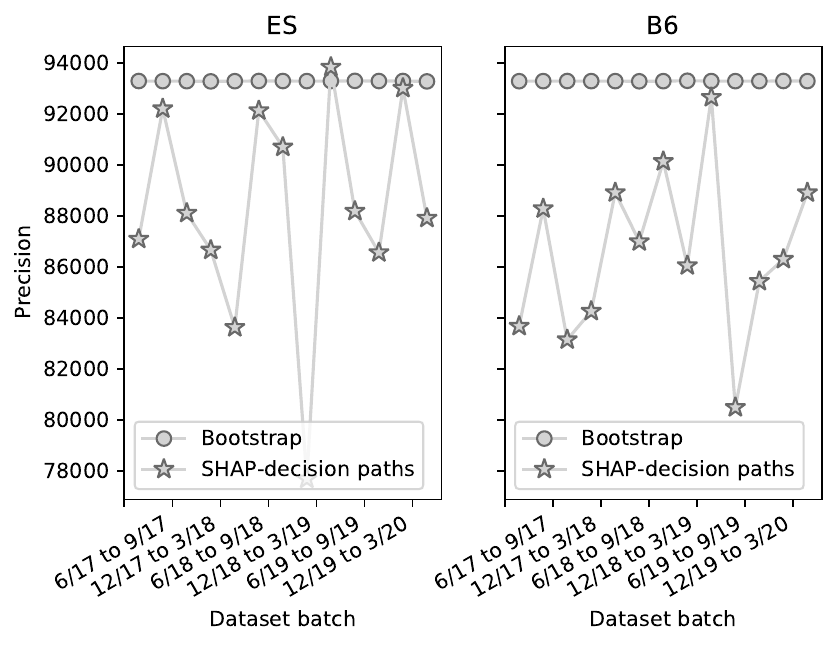}
    \caption{Footrule distances between ranks of the feature interactions for SHAP and decision path-based methods for S\&P E-mini and British Pound futures instruments.}
    \label{fig:FeatInteracBootstrapDistsSuppl}
\end{figure}



\end{document}